\documentclass[10pt,journal,compsoc]{IEEEtran}
\usepackage{amsmath,amsfonts}
\usepackage{algorithmic}
\usepackage{algorithm}
\usepackage{array}
\usepackage[caption=false,font=normalsize,labelfont=sf,textfont=sf]{subfig}
\usepackage{textcomp}
\usepackage{stfloats}
\usepackage{url}
\usepackage{verbatim}
\usepackage{graphicx}
\usepackage{cite}
\usepackage{hyperref}
\usepackage{cleveref}
\usepackage{booktabs}
\usepackage{multirow}
\usepackage{makecell}
\usepackage{mdwlist}
\usepackage{float}
\usepackage{subfig}
\usepackage{color,xcolor}
\usepackage{colortbl}
\usepackage{fancyhdr}
\usepackage{marvosym}
\usepackage{amssymb}
\usepackage{ragged2e}
\usepackage{tikz}
\usepackage{forest}
\usepackage{amsmath}
\hypersetup{
hidelinks,
colorlinks=true,
linkcolor=red,
citecolor=blue
}

\definecolor{best}{RGB}{242, 158, 156}
\definecolor{second}{RGB}{248, 206, 160}
\definecolor{third}{RGB}{255, 254, 166}

\begin{document}

\title{Attention in Diffusion Model: \\ A Survey}

\author{Litao Hua, Fan Liu, Jie Su, Xingyu Miao, Zizhou Ouyang, Zeyu Wang, Runze Hu, Zhenyu Wen,\\ Bing Zhai, Yang Long, Haoran Duan, Yuan Zhou

\IEEEcompsocitemizethanks{
\IEEEcompsocthanksitem Yuan Zhou, Litao Hua and Fan Liu are with the School of Artificial Intelligence, Nanjing University of Information Science and Technology, China. E-mail: zhouyuan@nuist.edu.cn; 202412621441@nuist.edu.cn; lfsss123123@gmail.com.
\IEEEcompsocthanksitem Haoran Duan is with Department of Automation, Tsinghua University. E-mail: haoranduan28@gmail.com.
\IEEEcompsocthanksitem Zhenyu Wen and Jie Su are with the Institute of Cyberspace Security and College of Information Engineering, Zhejiang University of Technology, Hangzhou 310023, China. (E-mail: zhenyuwen@zjut.edu.cn, jieamsu@gmail.com)
\IEEEcompsocthanksitem Zizhou Ouyang is with University of Edinburgh, UK. E-mail:zizhou.ouyang@ed.ac.uk.
\IEEEcompsocthanksitem Runze Hu is with the Beijing Institute of Technology, E-mail: hrzlpk2015@gmail.com.
\IEEEcompsocthanksitem Zeyu Wang is with College of Computer Science and Engineering, Dalian Minzu University. E-mail:20231578@dlnu.edu.cn.
\IEEEcompsocthanksitem Yang Long and Xingyu Miao are with the Department of Computer Science, Durham University, UK (e-mail: yang.long@durham.ac.uk, miaoxy97@163.com)
\IEEEcompsocthanksitem Bing Zhai is with the Department of Computer and Information Sciences, Northumbria University, Newcastle Upon Tyne, UK. E-mail: bing.zhai@northumbria.ac.uk.
\IEEEcompsocthanksitem Litao Hua and Fan Liu have equal contribution.
\IEEEcompsocthanksitem Yuan Zhou is the corresponding author.
}
\thanks{Manuscript submitted April 1, 2025;}}

\markboth{}
{Shell \MakeLowercase{\textit{et al.}}: Bare Demo of IEEEtran.cls for Computer Society Journals}


\IEEEtitleabstractindextext{%
\begin{abstract}
\justifying
Attention mechanisms have become a foundational component in diffusion models, significantly influencing their capacity across a wide range of generative and discriminative tasks. This paper presents a comprehensive survey of attention within diffusion models, systematically analysing its roles, design patterns, and operations across different modalities and tasks. We propose a unified taxonomy that categorises attention-related modifications into parts according to the structural components they affect, offering a clear lens through which to understand their functional diversity. In addition to reviewing architectural innovations, we examine how attention mechanisms contribute to performance improvements in diverse applications. We also identify current limitations and underexplored areas, and outline potential directions for future research. Our study provides valuable insights into the evolving landscape of diffusion models, with a particular focus on the integrative and ubiquitous role of attention.

\end{abstract}

}

\maketitle

\section{Introduction}
\IEEEPARstart{D}{iffusion} models \cite{ho2020denoising,song2020denoising,rombach2022high} have emerged as a powerful tool in deep learning, gaining attention for their ability to model complex data distributions. These models have proven particularly effective in both generative and discriminative tasks, although their application is predominantly seen in generative tasks. In recent years, diffusion models have found widespread use across various industries, ranging from healthcare to entertainment, where they contribute to advancements in data synthesis, anomaly detection, and optimization problems. In the realm of academic research, diffusion models have made significant strides, especially in the fields of natural language processing\cite{hu2020introductory} and computer vision\cite{hassanin2024visual}. The ability to generate realistic and coherent data has spurred innovations in multimodal generation tasks, such as text-to-image generation \cite{song2020denoising,huang2023kv,shi2024instantbooth,mou2023dragondiffusion}, style transfer \cite{deng2023z,yang2023zero}, image editing \cite{cao2023masactrl,nam2024dreammatcher,hertz2022prompt}, text-to-video generation \cite{zhou2022magicvideo,hong2022cogvideo,wu2023tune} and 3D generation \cite{wang2024gaussianeditor,liu2024dynvideo,yang2024dreamcomposer,pandey2024diffusion,chen2024gaussianeditor}. These applications have not only enhanced the creative capabilities of artificial intelligence but have also paved the way for new methodologies in deep learning.

\begin{figure}[!t]
\centering
\includegraphics[width=0.5\textwidth]{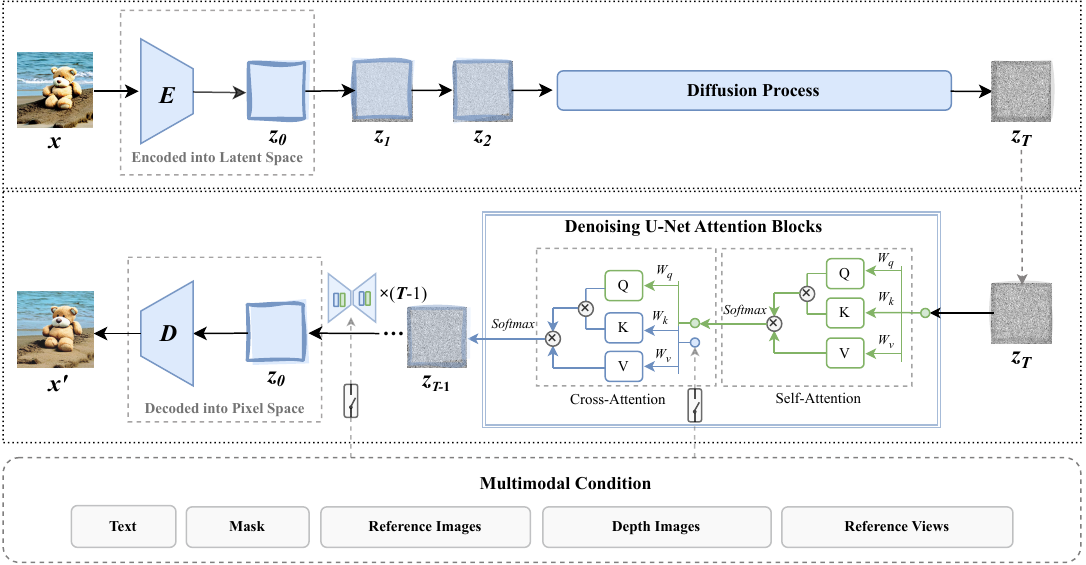}
\caption{\label{fig:overall pipeline}A typical pipeline of diffusion models, highlighting the attention mechanism for clarity. The pipeline consists of two stages: diffusion and denoising. Initially, the original image $x$ is encoded and gradually noised into $z_T$. Then, starting from $z_T$, the denoising U-Net, utilizing both cross-attention and self-attention, removes noise and reconstructs the image $x'$. Notably, the attention blocks within U-Net are presented in detail, illustrating how cross-attention and self-attention are implemented and interact. This detailed representation is crucial for understanding the model's internal workings, especially regarding the attention mechanisms.}
\end{figure}

The core pipeline of a diffusion model, shown in Fig.~\ref{fig:overall pipeline} involves the gradual transformation of noise into structured data through a series of iterative denoising steps\cite{ho2020denoising, song2020denoising, rombach2022high}. These models typically rely on architectures such as U-Net, which predict the denoised data at each step. While diffusion models have proven effective across various tasks, including both generative and discriminative tasks, a key challenge lies in capturing and maintaining the complex relationships between features and their interactions. These models must not only learn dynamic patterns that evolve over time but also ensure the controlled generation of outputs and improve prediction accuracy. To achieve this, an efficient method of dynamically weighting and aligning features is required, whether for image synthesis, segmentation, or other tasks. This is where attention mechanisms become indispensable\cite{ho2020denoising,song2020denoising}. Attention mechanisms allow the model to selectively prioritize and dynamically adjust the importance of features, enabling it to focus on the most relevant parts of the input. By dynamically attending to varying parts of the input at each step, the model can learn intricate dependencies across features, improving both the quality, accuracy and interpretability of the results. This ability to focus on critical parts of the data enables the model to capture both local details and broader contextual information\cite{guo2022attention,vaswani2017attention}.
In generative tasks, such as text-to-image generation, attention mechanisms are crucial to align textual and visual representations\cite{cao2023masactrl,hertz2022prompt}. Attention enables the model to focus on key attributes in the text and match them to relevant visual features dynamically. Unlike traditional feature extraction methods, attention mechanisms provide flexibility in how different parts of the input are weighted, allowing for a more nuanced interpretation of the text and ensuring the generated image aligns with the intended description\cite{rombach2022high}.
In discriminative tasks, such as semantic segmentation\cite{asiedu2022decoder}, attention plays a pivotal role in enhancing the model’s ability to focus on specific regions of an image that are critical for classification. However, in contrast to generative tasks, the focus here is not to produce new content but to refine the model’s understanding of the input’s structure\cite{ji2023ddp}. Attention allows the model to selectively refine its predictions by concentrating on regions that contain key features for pixel-wise classification. When segmenting an object from its background, attention ensures that fine details, such as object boundaries or textures, are more accurately delineated\cite{wu2023diffumask,pnvr2023ld}. This enables more accurate and contextually aware segmentation, enhancing the overall predictive capability of the model.

\begin{figure}[!t]
\centering
\includegraphics[width=0.5\textwidth]{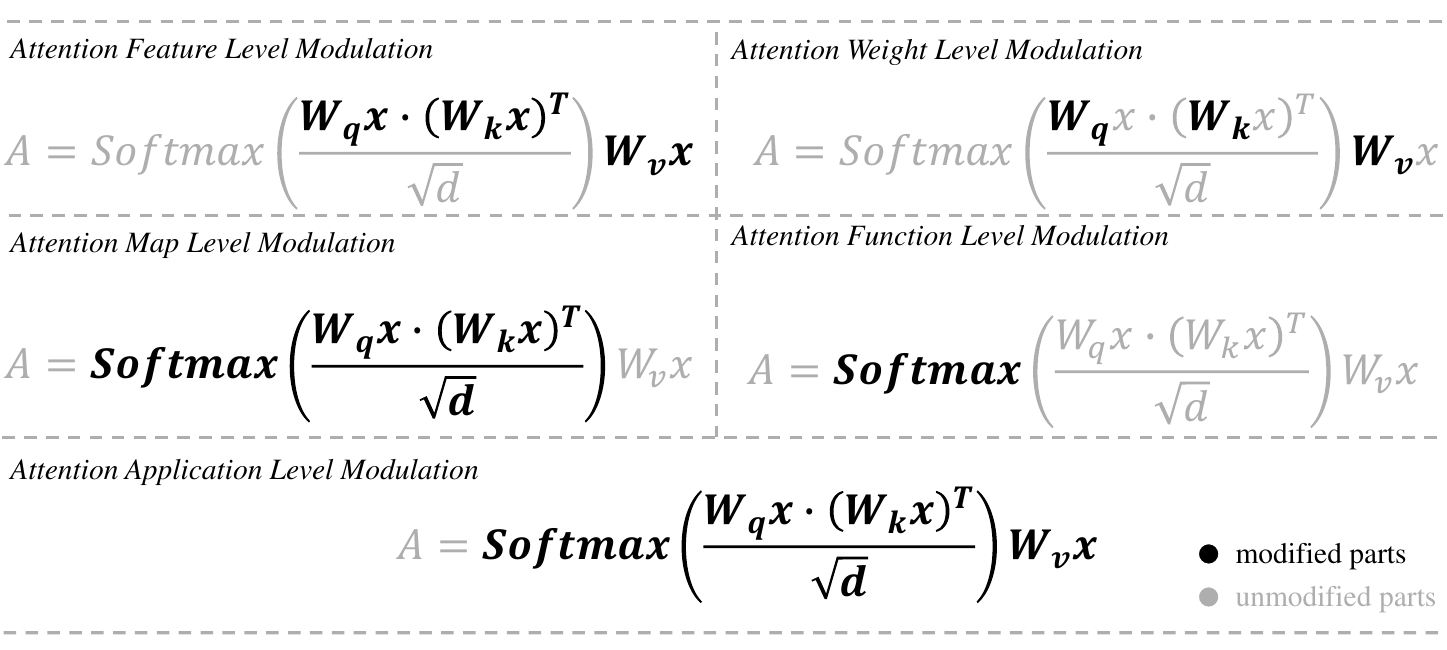}
\caption{\label{fig:idea}An illustration of the method to identify components of attention in diffusion model. $W_q$, $W_k$ and $W_v$ represent weight matrix for the query, key and value, respectively. $x$ stands for the input and $d$ is the scaling factor. We categorized the attention modifications into 5 levels based on the changes made to different components of attention. In each level, the modified parts are highlighted in black, while the unmodified parts are shown in gray.}
\end{figure}

\begin{figure*}[ht]
\centering
\includegraphics[width=1.0\textwidth]{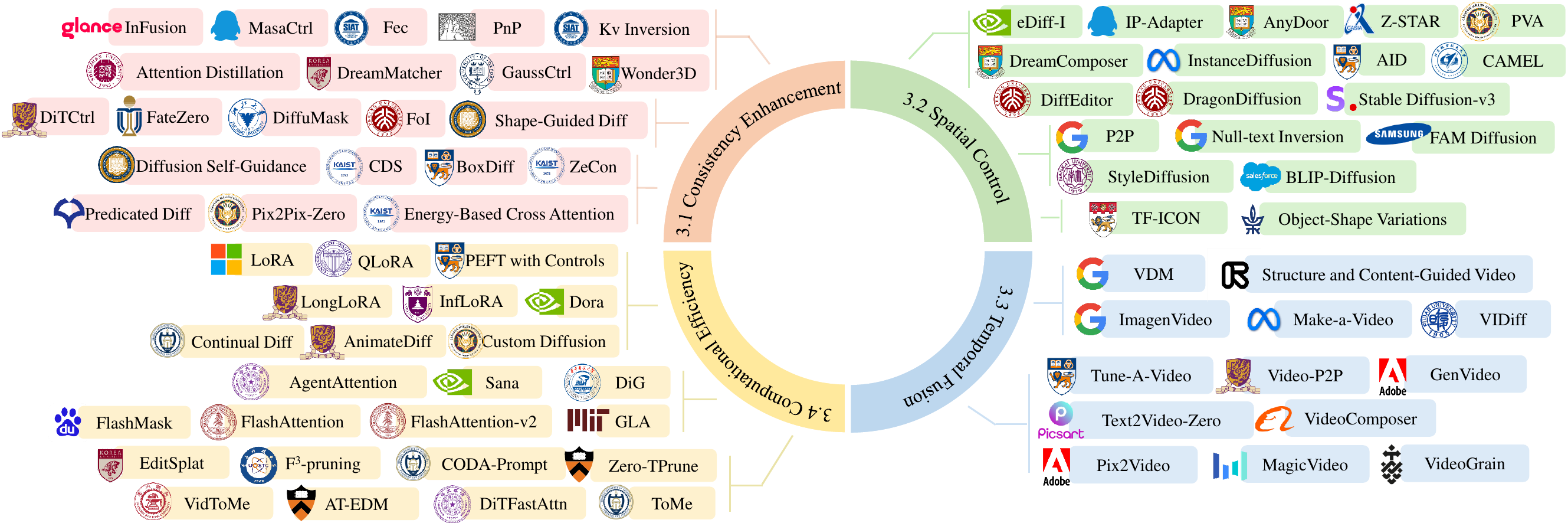}
\caption{\label{fig:category}Taxonomy of attention methods in diffusion models.}
\end{figure*}

Despite the remarkable success of attention mechanisms in diffusion models across various tasks, several challenges remain when it comes to feature extraction and cross-modal alignment. Issues such as inconsistency\cite{cao2023masactrl,tumanyan2023plug,nam2024dreammatcher}, lack of precise control\cite{hertz2022prompt,mou2024diffeditor,chen2024anydoor}, difficulty in integrating temporal features\cite{ho2022video,liu2024video}, and low computational efficiency\cite{han2024agent,hu2021lora,dao2022flashattention} still exist. Given the pivotal role of attention, many researchers have made significant contributions to modifying attention mechanisms in diffusion models to address these issues, thereby advancing the field. However, these noteworthy works lack a comprehensive and systematic review.
To address this gap, our paper systematically classifies existing methods along two key dimensions: the specific subproblems they target and their respective applications. We provide a thorough analysis of the similarities, differences, strengths, and limitations of each approach. In doing so, we offer a clear and structured overview of the evolving landscape of attention-enhanced diffusion models and present insights into potential directions for future advancements. Different from previous surveys\cite{cao2024survey, yang2023diffusion, croitoru2023diffusion, ulhaq2022efficient}, our work deconstructs the components of attention in diffusion models. This allows for better classification and a deeper understanding of how attention works at different stages and in different modalities. Based on the modified and unmodified components, we classify attention modification methods into five levels. The taxonomy of attention methods is shown in Fig.~\ref{fig:category}. The main contributions of this paper are as follows: \begin{itemize} \item A comprehensive and systematic taxonomy of attention mechanisms in multimodal diffusion models, highlighting the different roles and modulation strategies of attention across various stages of the diffusion process. \item A thorough exploration of the diverse application scenarios of multimodal diffusion models, offering valuable insights into their practical uses across different domains. \item A critical identification of the current challenges and limitations in attention-based diffusion models, along with proposed strategies for overcoming these issues, thus guiding future research directions in this rapidly developing field. \end{itemize}

The rest of this paper is organized as follows. We give a self-contained and brief introduction to the basic diffusion model and canonical attention mechanism in Section~\ref{background}. Section~\ref{sec:Roles and Modulation Methodologies of Attention in Diffusion Models} reviews and classifies the existing attention methods into 4 categories. Section~\ref{sec:Related Applications} provides a summary of the applications of multimodal generation using attention mechanisms. Finally, Section~\ref{sec:Challenges and Directions} highlights the limitations of current approaches and outlines promising directions for future research.

\begin{figure}[!t]
\centering
\includegraphics[width=0.5\textwidth]{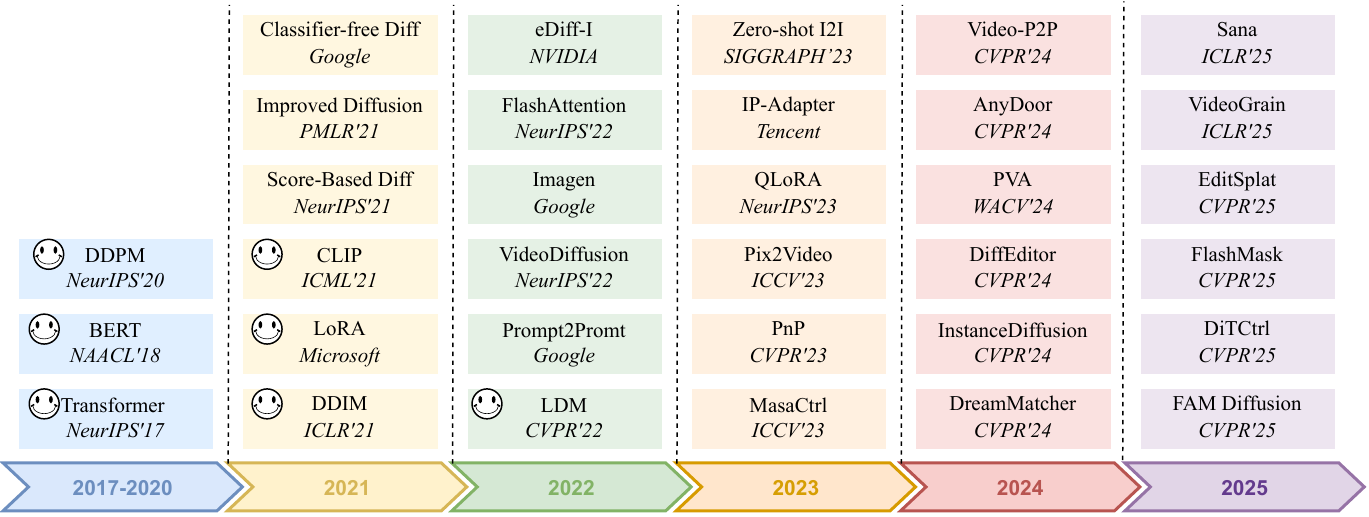}
\caption{\label{fig:timeline}The timeline of the development of attention related methods and diffusion models. The boxes indicate representative works. The boxes marked with a smile symbol represent the foundation models in this field.}
\end{figure}

\section{Background}
\label{background}
\subsection{Other Surveys}

In this section, we briefly compare our work with various existing surveys that have reviewed diffusion models and attention mechanisms. Two notable surveys \cite{guo2022attention, hassanin2024visual} focus on attention methods in deep neural networks, with an emphasis on their application in computer vision. These surveys primarily discuss recurrent neural network-based and Transformer-based models, whereas our study focuses on diffusion models, offering a distinct perspective.

More specialized surveys \cite{cao2024survey, yang2023diffusion, croitoru2023diffusion, ulhaq2022efficient} summarize the development of diffusion models, concentrating on diffusion sampling methods and architectural designs in vision applications. However, these works pay limited attention to the role of attention mechanisms within diffusion models. Yi Huang et al. \cite{huang2024diffusion} present a survey on diffusion models in image editing tasks. While their review mentions improved attention mechanisms within diffusion models, it is restricted to a single-modal task and offers only a superficial exploration. In contrast, our work provides a broader investigation and deeper analysis of the multimodal applications of attention mechanisms in diffusion models.

Additionally, unlike previous surveys, we introduce a novel taxonomy that categorizes various attention methods in diffusion models based on their roles and the modulation at different levels, which is shown in Fig.~\ref{fig:idea}. This classification allows for a comprehensive analysis of the interaction between attention mechanisms and diffusion models, highlighting when and where attention mechanisms play a critical role. By doing so, we move beyond treating attention mechanisms as merely supplementary components to other tasks, offering a more integrated perspective.

\subsection{Attention in Diffusion Models}
\subsubsection{Diffusion Models: Principals and Development}

In the domain of AI-Generated Content \cite{cao2023comprehensive,duan2025parameter}, diffusion models \cite{ho2020denoising,nichol2021improved,song2020denoising,rombach2022high} have led to remarkable advancements in generative tasks. The development timeline can refer to Fig.~\ref{fig:timeline}. These models operate by progressively adding noise to data in the forward process and subsequently learning to reverse this process. Specifically, Denoising Diffusion Probabilistic Models (DDPM)\cite{ho2020denoising} generate data samples by sampling an initial noise vector from a prior distribution and progressively denoising it into the desired data using a learnable reverse-time Markov chain.. Starting with a data sample $x_0$, a sequence of noisy samples $x_1$, $x_2$, $x_3$, …, $x_T$ is generated, where $T$ is the total number of time steps. The forward process can be defined as:
\begin{equation}
q(x_t \mid x_{t-1}) = \mathcal{N}(x_t; \sqrt{1-\beta_t} x_{t-1}, \beta_t \mathbf{I})
\end{equation}

For analytical convenience, $x_t$ can be sampled directly from the distribution of $x_0$. it can be rewritten as:
\begin{equation}
q(x_t \mid x_0) = \mathcal{N}(x_t; \sqrt{\bar{\alpha}_t} x_0, (1 - \bar{\alpha}_t) \mathbf{I})
\end{equation}
where $\beta_t$ is a variance schedule controlling the amount of noise added at each step $t$. $\mathcal{N}$ denotes a Gaussian distribution. $\mathbf{I}$ is the identity matrix and $\alpha_t = 1 - \beta_t$, $\bar{\alpha}_t = \prod_{i=1}^{t} \alpha_i$. These two operations play a critical role in controlling the noise schedule and regulating the variance of the process.

Starting from random noise, the reverse process iteratively refines it to generate data that aligns with the distribution of the original source. It's parameterized by a neural network $\theta$ to predict the noise added at each time step. The reverse process is modeled as:
\begin{equation}
p_\theta(x_{t-1} \mid x_t) = \mathcal{N}(x_{t-1}; \mu_\theta(x_t, t), \Sigma_\theta(x_t, t))
\end{equation}
where $\mu_\theta(x_t, t)$ and $\Sigma_\theta(x_t, t)$ are the mean and variance parameterized by a neural network.

The simplified training loss directly compares the true noise added in the forward process with the noise predicted by the model, which is defined as follows:
\begin{equation}
L(\theta) = \mathbb{E}_{x_0, t, \epsilon} \left[ \|\epsilon - \epsilon_\theta(x_t, t)\|^2 \right]
\end{equation}
where $\epsilon$ is the true noise added to the sample. $\epsilon_\theta(x_t, t)$ is the noise predicted by the model at time step $t$.

Based on DDPM, Denoising Diffusion Implicit Models (DDIM) \cite{song2020denoising} introduced a deterministic reverse diffusion process that skips random sampling, significantly accelerating the generation process. The DDIM sampling equation is as follows:
\begin{equation}
x_{t-1} = \sqrt{\alpha_{t-1}} \left( \frac{x_t - \sqrt{1 - \alpha_t} \epsilon_\theta(x_t, t)}{\sqrt{\alpha_t}} \right) + \sqrt{1 - \alpha_{t-1}} \epsilon_\theta(x_t, t)
\end{equation}

The first term “removes” part of the noise from $x_t$  and estimates  $x_{t-1}$ at the previous diffusion step. The second term adjusts the sample using the noise $\epsilon_\theta(x_t, t)$ predicted by the neural network, without introducing any additional random noise.

To further enhance computational efficiency, Latent Diffusion Model (LDM)\cite{rombach2022high} performs diffusion in latent space. Specifically, an autoencoder (e.g., a variational autoencoder (VAE) \cite{kingma2013auto}) first compresses the data samples into a lower-dimensional latent representation. A diffusion model is then applied in this latent space, and the latent variables are subsequently decoded back into the original data space.  This procedure significantly reduces computational costs compared to operating directly in high-dimensional pixel space.

In summary, DDPM is a diffusion model based on a random Markov chain, which provides high-quality generation results but suffers from slow sampling speed. DDIM improves efficiency by reducing the number of steps through a deterministic reverse process. LDM, on the other hand, performs diffusion in latent space, substantially lowering computational costs and making it better suited for high-resolution and complex scenes. As a result, most contemporary applications primarily adopt LDM, as it effectively balances efficiency and primarily adopt, particularly for large-scale, high-resolution tasks.

\begin{table*}[htbp]
\caption{Comprehensive categorization of attention mechanisms in diffusion model from multiple perspectives.}
\centering
\resizebox{\textwidth}{!}{
\begin{tabular}{|c|c|c|c|c|}
\hline
\textbf{Type} & \textbf{Method} & \textbf{Venue} & \textbf{Backbone} & \textbf{Modality} \\
\hline\hline  
\multirow{9}{*}{\makecell{Self-Attention Feature Injection \\ (Attention Feature Level)}} & MasaCtrl\cite{cao2023masactrl} & ICCV 2023 & Stable Diffusion-v1.4 \& Anything-v3 & Text \& Image \\
& Fec\cite{10424833} & ICML 2023 & Stable Diffusion & Text \& Image \\
& InFusion\cite{khandelwal2023infusion} & ICCV 2023 & Stable Diffusion-v1.5 & Video \& Text \\
& Kv Inversion\cite{huang2023kv} & PRCV 2023 & Anything-v3 & Text \& Image \\
& PnP\cite{tumanyan2023plug} & CVPR 2024 & Stable Diffusion & Text \& Image \\
& DreamMatcher\cite{nam2024dreammatcher} & CVPR 2024 & Stable Diffusion-v1.4 & Text \& Image \\
& Wonder3D\cite{Long_2024_CVPR} & CVPR 2024 & Stable Diffusion & 3D \& Image \\
& GaussCtrl\cite{gaussctrl2024} & ECCV 2024 & ControlNet & 3D \& Text \& Image \\
& Attention Distillation\cite{zhou2025attention} & CVPR 2025 & Stable Diffusion-v1.5 & Text \& Image \\
\hline\hline
\multirow{5}{*}{\makecell{Attention-based Mask Guidance\\(Attention Application Level)}} 
& DiffuMask \cite{wu2023diffumask} & ICCV 2023 & Stable Diffusion \& CLIP & Text \& Image \\
& FateZero \cite{qi2023fatezero} & ICCV 2023 & Stable Diffusion-v1.4 & Video \& Text \& Image \\
& FoI\cite{guo2024focus} & CVPR 2024 & Instructpix2pix \& CLIP \& GPT-4 & Text \& Image \\
& Shape-Guided Diffusion\cite{park2024shape} & WACV 2024 & Stable Diffusion & Text \& Image \\
& DiTCtrl\cite{cai2024ditctrl} & CVPR 2025 & Multimodal Diffusion Transformer & Video \& Text \& Image \\
\hline\hline
\multirow{7}{*}{\makecell{Attention Score-Driven Guidance\\(Attention Application Level)}} & Pix2Pix-Zero\cite{parmar2023zero} & SIGGRAPH 2023 & Stable Diffusion-v1.4 & Text \& Image \\
& BoxDiff\cite{xie2023boxdiff} & ICCV 2023 & Stable Diffusion & Text \& Image \\
& ZeCon\cite{yang2023zero} & ICCV 2023 & Unconditional Stable Diffusion \& CLIP & Image \\
& Diffusion Self-Guidance\cite{epstein2023diffusion} & NIPS 2023 & - & Text \& Image \\
& CDS\cite{nam2024contrastive} & CVPR 2024 & Stable Diffusion-v1.4 & Text \& Image \\
& Predicated Diffusion\cite{sueyoshi2024predicated} & CVPR 2024 & Stable Diffusion-v1.4 & Text \& Image \\
& Energy-Based Cross Attention\cite{park2024energy} & NIPS 2024 & Stable Diffusion \& CLIP & Text \& Image \\
\hline\hline
\multirow{12}{*}{\makecell{Conditional Alignment in Cross-Attention\\(Attention Feature Level)}} & eDiff-I\cite{balaji2022ediff} & arXiv-2022 & Stable Diffusion \& CLIP \& T5 & Text \& Image \\
& IP-Adapter\cite{ye2023ip} & arXiv-2023 & Stable Diffusion-v1.5 \& OpenCLIP ViT-H/14 & Text \& Image \\
& Z-STAR\cite{deng2023z} & arXiv-2023 & Stable Diffusion-v1.5 & Image \\
& DragonDiffusion\cite{mou2023dragondiffusion} & ICLR 2024 & Stable Diffusion-v1.5 & Text \& Image \\
& AnyDoor\cite{chen2024anydoor} & CVPR 2024 & Stable Diffusion \& DINOv2 & Text \& Image \\
& DiffEditor\cite{mou2024diffeditor} & CVPR 2024 & Stable Diffusion-v1.5 & Text \& Image \\
& DreamComposer\cite{yang2024dreamcomposer} & CVPR 2024 & Zero-1-to-3 & 3D \& Text \& Image \\
& InstanceDiffusion\cite{wang2024instancediffusion} & CVPR 2024 & Stable Diffusion \& BLIP-V2 \& Ground-SAM & Text \& Image \\
& CAMEL\cite{zhang2024camel} & CVPR 2024 & Stable Diffusion-v1.4 & Video \& Text \& Image\\
& PVA\cite{Xu_2024_WACV} & WACV 2024 & Latent Diffusion Inpainting & Text \& Image \\
& AID\cite{he2024aid} & NIPS 2024 & Stable Diffusion-v1.5 & Text \& Image \\
& Stable Diffusion-v3\cite{esser2024scaling} & arXiv-2024 & Stable Diffusion-v3 & Text \& Image \\
\hline\hline
\multirow{5}{*}{\makecell{Cross-Attention Map Control\\(Attention Map Level)}} & P2P\cite{hertz2022prompt} & ICLR 2023 & LDM \& Stable Diffusion & Text \& Image\\
& Null-text Inversion\cite{Mokady_2023_CVPR} & arXiv-2023 & Stable Diffusion & Text \& Image \\
& StyleDiffusion\cite{li2023stylediffusion} & arXiv-2023 & Stable Diffusion & Text \& Image \\
& BLIP-Diffusion\cite{li2024blip} & NIPS 2024 & LDM \& ControlNet & Text \& Image \\
& FAM Diffusion\cite{yang2024fam} & CVPR 2025 & Stable Diffusion XL & Image \\
\hline\hline
\multirow{2}{*}{\makecell{Selective Attention Map Composition\\(Attention Map Level)}} & Object-Shape Variations\cite{patashnik2023localizing} & ICCV 2023 & Stable Diffusion & Text \& Image \\
& TF-ICON\cite{lu2023tf} & ICCV 2023 & Stable Diffusion & Text \& Image \\
\hline\hline
\multirow{5}{*}{\makecell{Temporal Attention Injection\\(Attention Feature Level)}} & ImagenVideo\cite{ho2022imagen} & arXiv-2022 & DDPM & Video \& Text \& Image \\
& VDM\cite{ho2022video} & NIPS 2022 & DDIM & Video \& Text \& Image \\
& Make-a-Video\cite{singer2022make} & arXiv-2022 & - & Video \& Text \& Image \\
& Structure and Content-Guided Video\cite{Esser_2023_ICCV} & ICCV 2023 & LDM & Video \& Text \& Image \\
& VIDiff \cite{xing2023vidiff} & arXiv-2023 & Stable Diffusion-v1.5 & Video \& Text \& Image\\
\hline\hline
\multirow{9}{*}{\makecell{Spatio-Temporal Feature Alignment\\(Attention Feature Level)}} & MagicVideo\cite{zhou2022magicvideo} & arXiv-2023 & LDM \& VAE \& CLIP & Video \& Text \& Image \\
& VideoComposer\cite{NEURIPS2023_180f6184} & NIPS 2023 & LDM & Video \& Text \& Image \\
& Pix2Video\cite{ceylan2023pix2video} & ICCV 2023 & Stable Diffusion & Video \& Text \& Image \\
& Text2Video-Zero\cite{Khachatryan_2023_ICCV} & ICCV 2023 & Stable Diffusion-v1.5 & Video \& Text \& Image \\
& Tune-A-Video\cite{wu2023tune} & ICCV 2023 & Stable Diffusion & Video \& Text \& Image \\
& GenVideo\cite{harsha2024genvideo} & CVPR 2024 & Stable Diffusion-v2.1 & Video \& Text \& Image \\
& Video-P2P\cite{liu2024video} & CVPR 2024 & Stable Diffusion-v1.5 & Video \& Text \& Image \\
& VideoGrain\cite{yang2025videograin} & ICLR 2025 & Stable Diffusion-v1.5 & Video \& Text \& Image \\
\hline\hline
\multirow{3}{*}{\makecell{Linear Attention\\(Attention Function Level)}} & AgentAttention\cite{han2024agent} & ECCV 2024 & - & Text \& Image \\
& DiG\cite{zhu2024dig} & arXiv-2024 & Gated Linear Transformer \& DDPM & Image \\
& Sana\cite{xie2024sana} & ICLR 2025 & Diffusion Transformer & Text \& Image \\
\hline\hline
\multirow{4}{*}{\makecell{Chunk Attention\\(Attention Function Level)}} & FlashAttention\cite{dao2022flashattention} & arXiv-2023 & - & Text \& Image \\
& FlashAttention-v2\cite{dao2023flashattention} & NIPS 2023 & - & Text \& Image \\
& GLA\cite{yang2023gated} & arXiv-2024 & - & Text \\
& FlashMask\cite{wang2024flashmask} & ICLR 2025 & - & - \\
\hline\hline
\multirow{7}{*}{\makecell{LoRA based Finetuning\\(Attention Weight Level)}} & LoRA\cite{hu2021lora} & arXiv-2021 & - & - \\
& QLoRA\cite{dettmers2024qlora} & NIPS 2023 & - & - \\
& LongLoRA\cite{chen2023longlora} & ICLR 2024 & - & - \\
& InfLoRA\cite{liang2024inflora} & CVPR 2024 & - & - \\
& Dora\cite{liu2024dora} & ICML 2024 & - & Video \& Text \& Image \\
&PEFT with Controls\cite{zhangparameter} & ICML 2024 & ViT & Image \\
&AnimateDiff\cite{guo2023animatediff} & ICLR 2024 & Stable Diffusion-v1.5 & Text \& Image \\
\hline\hline
\multirow{2}{*}{\makecell{Selective Finetuning\\(Attention Weight Level)}} & Custom Diffusion\cite{kumari2023multi} & CVPR 2023 & Stable Diffusion & Text \& Image \\
& Continual Diffusion\cite{smith2024continual} & TMLR 2024 & Stable Diffusion & Text \& Image \\
\hline\hline
\multirow{8}{*}{\makecell{Attention-based Sparsification and Token Pruning\\(Attention Application Level)}} & CODA-Prompt\cite{smith2023coda} & CVPR 2023 & - & Image \\
& ToMe\cite{Bolya_2023_CVPR} & CVPR 2023 & Stable Diffusion & Image \\
& VidToMe\cite{li2024vidtome} & CVPR 2023 & Stable Diffusion-v1.5 & Video \& Text \& Image \\
& F$^3$-pruning\cite{su2024f3} & AAAI 2024 & - & Video \& Text \& Image \\
& DiTFastAttn\cite{yuan2024ditfastattn} & NIPS 2024 & Diffusion Transformers & Video \& Text \& Image \\
& Zero-TPrune\cite{wang2024zero} & CVPR 2024 & - & Image \\
& AT-EDM\cite{wang2024attention} & CVPR 2024 &  Stable Diffusion-XL & Text \& Image \\
& EditSplat\cite{in2024editsplat} & CVPR 2025 & InstructPix2Pix & 3D \& Text \& Image \\
\hline
\end{tabular}
}
\label{tab:Comprehensive Categorization Table}
\end{table*}

\subsubsection{Attention Mechanism: Principals and its relationship with diffusion models}

Multimodal generation tasks often face challenges such as inconsistency, difficulties in controlling fine details, insufficient temporal information, and high computational complexity. Traditional generative models, like Generative Adversarial Networks (GANs)\cite{yang2019diversity,karras2019style}, address these problems by leveraging the global feature representation within the latent space. The latent space serves as a bridge, capturing abstract feature relationships that enable consistency and control during generation.
In contrast, diffusion models rely on attention to maintain global consistency while enhancing spatial and temporal control. To explain this process, we first need to delve into the principles of the attention mechanism and then explore how it integrates with diffusion models to address key challenges in multimodal generative tasks, including consistency, spatial control, temporal fusion, and computational efficiency.

Attention \cite{duan2023dynamic,duan2024dual,duan2024wearable} is a core element of the human cognitive system, enabling individuals to selectively filter and focus on pertinent information from a multitude of sensory inputs. Inspired by this cognitive process, computer scientists have developed attention mechanisms that replicate this ability, amplifying relevant data features while disregarding extraneous elements. In the traditional attention mechanism \cite{vaswani2017attention,ding2021repvgg}, the attention map is obtained by computing the cosine similarity between a given query and key, followed by a normalization process. These attention maps are then used to weight and sum elements of the input sequence, generating an attention-based output representation, as expressed in Eq. \ref{eq:attention}. 
\begin{equation}\label{eq:attention}
\text{Attention} = \text{softmax}(\frac{Q\cdot K^T}{\sqrt{d}})\cdot V 
\end{equation}
where $Q$, $K$ and $V$ stand for the query, key and value respectively. $d$ is a scaling factor. This output can be fed into subsequent processing stages, allowing the model to more effectively capture task-relevant information from the input data, thereby improving the model's overall performance and efficiency. The sources of Q, K, and V can vary depending on the task requirements. In self-attention, Q, K, and V come from the same sequence, while in cross-attention, they come from different sequences.

Attention mechanisms, particularly self-attention and cross-attention, play a crucial role in diffusion models. The stepwise generation process in diffusion models is complex. Each step gradually denoises the data to approach the desired output. At the same time, the model adapts to changing input conditions and data characteristics. Attention mechanisms, especially self-attention and cross-attention, guide this process. The backbone architecture of diffusion models commonly employs the U-Net framework. Attention mechanisms are integrated into the middle and higher levels of the encoder and decoder within U-Net. They ensure both progressive refinement and dynamic adaptation. Self-attention is adept at modeling the spatial dependencies within modalities of the input. By computing global correlations among features, self-attention ensures that the generation process maintains global consistency while simultaneously enhancing the semantic integrity of the generated data. Cross-attention, on the other hand, focuses on feature mapping and alignment between modalities.

By incorporating the attention mechanism, diffusion models can enhance generative capabilities in several ways. First, attention mechanisms can address the consistency problem by ensuring that the generated output aligns with the input conditions, which is crucial in generative tasks. In terms of spatial control, attention helps the model capture local features of the image during generation and adjusts the weighting between different parts of the image, allowing for precise spatial detail control. Regarding temporal fusion, attention mechanisms can help by combining information from different time steps, ensuring smooth transitions across the generation process and improving the stability of the model. Lastly, although attention mechanisms typically introduce higher computational complexity, more efficient variants, such as sparse attention, have been introduced to maintain high-quality generation while improving computational efficiency.

\section{Roles and Modulation Methodologies of Attention in Diffusion Models}
\label{sec:Roles and Modulation Methodologies of Attention in Diffusion Models}
This section systematically classifies and summarizes existing attention mechanisms in multimodal diffusion models from a methodological perspective. Multimodal diffusion models represent a significant frontier in generative model research, and the evolution of attention methods reflects key technical trends and conceptual innovations in the field. By adopting a methodological viewpoint, this section aims to systematically organize the design principles, optimization techniques, and novel contributions of various models.

Despite differences in specific implementations across models, attention mechanisms share commonalities at the methodological level, such as masking mechanisms, attention control, and the utilization of latent spaces. Summarizing these methods allows us to uncover these shared characteristics and distinctions, providing insights into the strengths and limitations of existing approaches. We identify the limitations of existing research and highlighted key areas for future exploration and necessary improvements.

To provide a comprehensive understanding, this chapter deconstructs the attention computation process into its constituent parts and classifies existing methods by analyzing how each part of attention is modulated, which is shown in Fig.~\ref{fig:idea}.  This classification clearly illustrates the stages at which different attention modulation techniques take effect, offering valuable insights into their roles within the overall process. More details can refer to Table.~\ref{tab:Comprehensive Categorization Table}.

\subsection{Consistency Enhancement}
\label{sec:Consistency Enhancement}
\begin{figure*}[ht]
\centering
\includegraphics[width=1.0\textwidth]{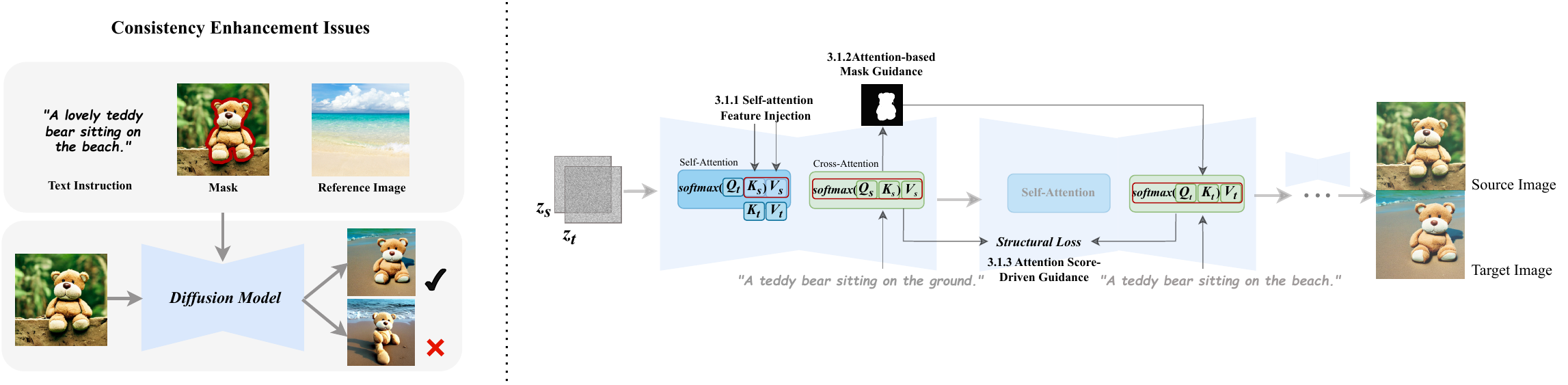}
\caption{\label{fig:class1}An illustration of a typical architecture of consistency enhancement. The left side of the figure illustrates the consistency issue, while the right side shows the method of modifying attention to maintain consistency. $Q_s$, $K_s$ and $V_s$ originate from the source image or text. $Q_t$, $K_t$ and $V_t$ come from the target image or text. The modulated components of attention are highlighted with red boxes.}
\end{figure*}

Consistency enhancement is a crucial objective in diffusion models, especially when dealing with tasks like editing, where maintaining coherent visual structures across modified and unmodified regions is essential\cite{cao2023masactrl}. The typical pipeline for editing tasks starts with selecting the content to modify. A generative model or editing tool like diffusion model is then used to process and alter the chosen areas, while ensuring that the changes blend naturally with the original content. One of the key challenges in diffusion models is ensuring that the generated outputs remain consistent throughout the denoising process, particularly in multimodal settings. To address this issue, several attention mechanisms have been developed to improve the consistency of the generated content. A typical pipeline of methods mentioned in this section can all refer to Fig.~\ref{fig:class1}.

\subsubsection{Self-Attention Feature Injection}
\label{sec:Self-Attention Feature Injection}
Self-attention feature injection\cite{10424833,cao2023masactrl,tumanyan2023plug,nam2024dreammatcher,huang2023kv,khandelwal2023infusion,gaussctrl2024,Long_2024_CVPR,zhou2025attention} focuses on selectively fuse features from the sources images within the self-attention layer of U-net to achieve consistency. In the standard self-attention mechanism, the query $Q$, key $K$ and value $V$ each focus on the similar information derived from the same input and are unable to focus on different aspects of the same input. For example, in text prompt-based image editing, it is often necessary to focus on the edited regions while keeping the unedited parts unchanged, a requirement that traditional self-attention mechanisms cannot fully meet. By employing a cross-attention-like mechanism to leverage features from the source image's reconstruction diffusion pipeline into the target image's denoising process, this method preserves unedited concepts, amplifies edited elements, and suppresses removed aspects within the editing diffusion pipeline, thereby reducing inconsistencies. The common pipeline is illustrated in Fig.~\ref{fig:class1}. Different methods replace different features. Attention Distillation\cite{zhou2025attention}, Fec\cite{10424833}, MasaCtrl\cite{cao2023masactrl}, Kv Inversion\cite{huang2023kv}, Infusion\cite{khandelwal2023infusion}, GaussCtrl\cite{gaussctrl2024} and Wonder3D\cite{Long_2024_CVPR} emphasizes modifying $K$ and $V$ in the decoder's attention layers, whereas PnP\cite{tumanyan2023plug} focuses more on the replacement of $Q$ and $K$. In DreamMatcher\cite{nam2024dreammatcher}, a warp operation is performed before replacing $V$ to establish semantic correspondence between the reference and target. While these approaches all aim to enhance consistency through attention modification, their applicability varies depending on the specific editing task. $Q$ and $K$ encode structural features and control the spatial arrangement of image elements. $V$ captures appearance features, such as colors, textures, and shapes, and assigns them to the corresponding image elements\cite{nam2024dreammatcher}. This distinction leads to the different strengths of each method in various tasks. MasaCtrl excels in action editing by modifying $K$ and $V$, ensuring structural consistency while allowing controlled changes in action. Rather than directly substituting $K$ and $V$, Attention Distillation\cite{zhou2025attention} leverages a teacher-student framework, where the K and V from the target image serve as supervisory signals to guide the learning of corresponding representations. PnP focuses on manipulating $Q$ and $K$ to preserve the structure, making it particularly effective in object editing. DreamMatcher specializes in scene editing, using a warp operation before replacing $V$ to align the appearance features between the reference and target. This ensures semantic and structural consistency in large-scale scene edits. While these methods perform well within their specific domains, they lack a unified framework for broader tasks. Future research could integrate the strengths of these methods into a more versatile editing approach, suitable for different tasks.

\subsubsection{Attention-based Mask Guidance}
\label{sec:Attention-based Masking and Guidance}
Masks are commonly used in editing and inpainting tasks to address the problem where the edited object can easily be confused with the background. Cross-attention maps associated with the prompts contain most of the shape and structure information. This information not only helps distinguish between foreground and background, but also plays a crucial role in locating regions of interest. The region of interest (ROI) associated with the prompts can be extracted using a mask derived from analyzing the cross-attention maps, which separate the ROI and background information to improve consistency, as demonstrated by FoI\cite{guo2024focus}, MasaCtrl\cite{cao2023masactrl}, DiTCtrl\cite{cai2024ditctrl}, DiffuMask \cite{wu2023diffumask} and Object-Shape Variations\cite{patashnik2023localizing}. MasaCtrl and Object-Shape Variations use the extracted mask to restrict the ROI in the target image's denoising process, allowing it to query content information only from the corresponding ROI region in the original image. Additionally, both the ROI and background regions query content from their respective restricted areas in the source image, rather than from all features. In contrast, FoI\cite{guo2024focus} focuses on adaptively applying this mask across each cross-attention layer. Another mask-based strategy is to constrain cross-attention maps with masks to locate the spatial region. Shape-Guided Diffusion\cite{park2024shape} infers the object mask from the source prompt as an input to both the self-attention and cross-attention layers. Constrained either by the object mask or its inverted counterpart, it produces a novel attention map called inside-outside attention. DiTCtrl\cite{cai2024ditctrl} generates masks by averaging relevant parts of the 3D full attention maps in multimodal Diffusion Transformers based on given object tokens. These masks are then used to guide attention fusion across different prompts, enabling precise semantic control. This approach ensures consistent object semantics and coherent motion in multi-prompt video generation. The common issue with these methods is their over-reliance on precise mask extraction. The accuracy of the mask extraction directly affects the distinction between foreground and background, as well as the quality of the model's generation. If the mask is not precise enough, it may result in unclear separation between foreground and background, causing artifacts or inconsistencies. It can also lead to difficulties when the model processes objects with complex shapes or rich details. FateZero\cite{qi2023fatezero} not only integrates spatio-temporal self-attention and cross-attention during DDIM inversion, but also leverages attention fusion and binary masks derived from cross-attention to enhance shape controllability while preserving temporal consistency. However, it struggles with layout preservation when performing local object editing. Overall, these methods still have room for improvement in terms of accuracy and adaptability to complex objects.

\begin{figure*}[ht]
\centering
\includegraphics[width=1.0\textwidth]{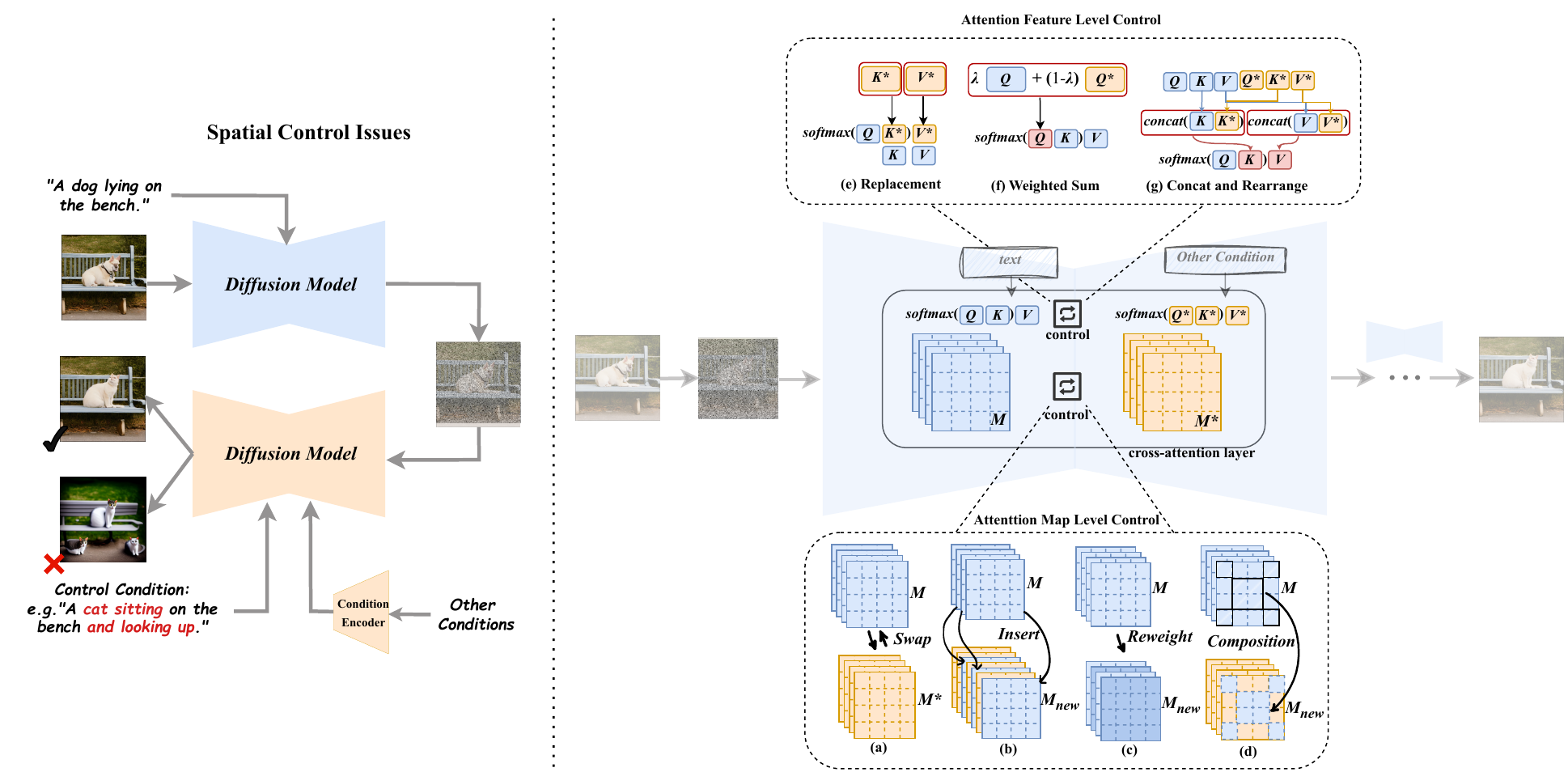}
\caption{\label{fig:class2} The figure illustrates a common pipeline: spatial control issue on the left and attention-based modification methods on the right. $Q$, $K$, $V$ with and without an asterisk (*) represent features obtained under two different conditions, respectively. $M$ and $M^*$ represent attention maps obtained under two different conditions. $M_{new}$ stands for the modified new attention map.}
\end{figure*}

\subsubsection{Attention Score-Driven Guidance}
The attention score guidance method \cite{nam2024contrastive,yang2023zero,sueyoshi2024predicated,epstein2023diffusion,park2024energy,parmar2023zero,xie2023boxdiff,li2023stylediffusion} utilizes the feature maps generated by the attention layer of the decoder in diffusion models to construct a loss or constraints, ensuring consistency throughout the generation process. Hyelin Nam $et$ $al.$ introduced Contrastive Denoising Score (CDS) \cite{nam2024contrastive}, which leverages the rich spatial information embedded in the self-attention features of LDM to compute the Contrastive Unpaired Translation (CUT) loss \cite{park2020contrastive}. ZeCon loss \cite{yang2023zero} has been proposed for image style transfer, maintaining semantic consistency between the reverse-sampled denoised image and the original, while preserving content information. Similarly, Predicated Diffusion \cite{sueyoshi2024predicated} derives a logic-based loss function from attention maps. Diffusion Self-Guidance\cite{epstein2023diffusion} introduces a self-guidance strategy, which extracts a set of properties from softmax-normalized attention matrices and activations, enabling control over generated images by adding guidance terms to the original loss function. Energy-Based Models (EBMs) \cite{park2024energy}, focusing on the cross-attention space of a time-dependent denoising autoencoder, minimize a specially designed energy function to correct semantic misalignment. Pix2Pix-Zero \cite{parmar2023zero} and StyleDiffusion \cite{li2023stylediffusion} employ an L2 loss to encourage the cross-attention maps of the source image to align with those of the edited versions. BoxDiff \cite{xie2023boxdiff}, under specific box conditions, calculates inner-box, outer-box, and corner constraints to guide image generation. Many of these methods\cite{nam2024contrastive,sueyoshi2024predicated,yang2023zero,epstein2023diffusion} rely on predefined structures, such as masks, logic-based cues, or L2 loss, which can constrain the model’s flexibility in handling more diverse or creative tasks. They often focus on maintaining consistency at the cost of introducing too much rigidity, leading to less diverse and potentially less realistic outputs. Some approaches, such as BoxDiff\cite{xie2023boxdiff}, are highly specialized and are more effective in constrained environments (e.g., images with defined boundaries). However, they may not generalize well to more dynamic or unconstrained scenes, limiting their applicability in diverse real-world scenarios.

\subsection{Spatial Control}
\label{sec:Spatial Control}
Spatial control is essential in diffusion models for managing the relationships between different regions of an image or across modalities. Attention mechanisms enable precise spatial focus, ensuring that the generated content aligns correctly with the intended target. This is particularly important in tasks like image-to-image translation or text-to-image generation, maintaining spatial coherence is essential for high-quality results.. Current methods primarily focus on refining cross-attention to achieve better spatial control. The common pipeline of these methods in this section can refer to Fig.~\ref{fig:class2}.

\subsubsection{Conditional Alignment in Cross-Attention}
\label{sec:Conditional Alignment in Cross-Attention}
Conditional alignment in cross-attention\cite{mou2024diffeditor,Xu_2024_WACV,ye2023ip,deng2023z,chen2024anydoor,balaji2022ediff,wang2024instancediffusion,yang2024dreamcomposer,he2024aid,mou2023dragondiffusion} aims to incorporate the query $Q$, key $K$, and value $V$ computed from various data types into the cross-attention layer, enabling the generation of content that satisfies specific conditions through the manipulation of these components. In condition-driven generation tasks, this method provides the possibility to inject different conditions, where distinct $Q$, $K$, and $V$ record the different feature information. This method can refer to Fig.~\ref{fig:class2}(e)(f)(g). Cross-Attention Feature Injection typically employs the following strategies:
a) Replace one or more of the $Q$, $K$, or $V$ features in the cross-attention layer with those obtained under different conditions\cite{chen2024anydoor,zhang2024camel}. This approach is effective when you want to inject specific conditions into the attention mechanism to guide the generation in a straightforward way. However, replacing features in cross-attention may risk discarding important cross-modal information, which could lead to inconsistency or loss of context when the original features are replaced too aggressively. b) Perform a weighted sum of attention maps to create a novel attention map from different conditions\cite{mou2024diffeditor,ye2023ip,gaussctrl2024,he2024aid}. This method is more flexible in terms of integrating multiple conditions without directly replacing features, allowing for a smoother blending of information from various inputs. While it enables more controlled generation, the challenge lies in determining optimal weight distributions for the different conditions, as improper weighting could lead to dominant or conflicting features, affecting the quality of the output. c) Rearrange and concatenate $Q$, $K$, and $V$ obtained under different conditions to generate a new attention map\cite{Xu_2024_WACV,deng2023z,balaji2022ediff,wang2024instancediffusion,mou2023dragondiffusion}. This approach integrates multiple conditions without needing to adjust weight hyperparameters, enabling the diffusion model to generate more stable and high-quality content. However, the concatenation process could lead to high-dimensional attention maps that may be computationally expensive
Unlike the attention feature injection mechanism mentioned above, each of $Q$, $K$, and $V$ is multimodal in this method. Stable Diffusion v3 \cite{esser2024scaling} introduces a multimodal feature fusion attention mechanism. Specifically, this approach maps the image patch embeddings and text embeddings, integrating both modalities into $Q$, $K$, and $V$ for the attention operation. This modulated attention allows the model to effectively fuse information from both text and images while maintaining the distinct characteristics of each modality.

\begin{figure*}[ht]
\centering
\includegraphics[width=1.0\textwidth]{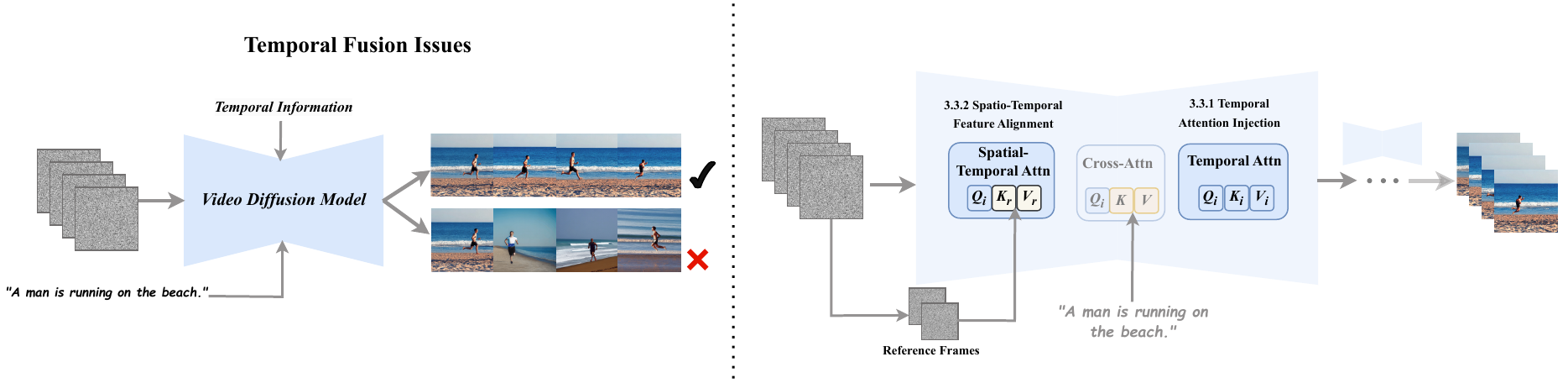}
\caption{\label{fig:class3} An illustration of the temporal issue (left) and a pipeline for temporal fusion methods with attention modification (right). $Q_i$, $K_i$ and $V_i$ originate from the $i$th frame of a video. $K_r$ and $V_r$ represent features from selected reference frames.}
\end{figure*}

\subsubsection{Cross-Attention Map Control}
\label{sec:Cross-Attention Map Control}

Cross-attention map control focuses on altering or influencing the softmax-normalized attention maps, which are defined as $Softmax(Q \cdot K)$. Researchers have extensively explored the semantic impact of cross-attention or self-attention\cite{kwon2022diffusion,preechakul2022diffusion} to control the generation of high-quality content in images, which is shown in Fig.~\ref{fig:class2}(a)(b)(c).

Recently, attention map control has emerged as one of the most effective techniques for detailed image generation \cite{hertz2022prompt,liu2024video,Mokady_2023_CVPR,li2024blip}. By simply modifying the condition, the desired contents can be generated without the need for additional training. Prompt-to-Prompt (P2P) \cite{hertz2022prompt} introduced a purely text-based editing framework that pioneered the use of cross-attention map control. This mechanism ensures structural consistency between edited and source images, allowing for precise adjustments while preserving key visual elements. This paper discusses three common control methods for image generation: a) Word swap: In this method, the attention maps from the editing path are replaced by the corresponding maps from the source path, ensuring alignment between the modified and original content. b) Adding a new phrase: When new tokens are introduced into the prompt, their attention maps are systematically inserted into the original cross-attention maps along the editing path, allowing for the seamless integration of new elements. c) Attention re-weighting: P2P adjusts the cross-attention map of a specific token by scaling it with a parameter, thereby either amplifying or diminishing its influence on the generated image. Since its introduction, many methods have partially adopted or fully built upon P2P due to its effectiveness and efficiency. Video-P2P \cite{liu2024video} applies the word swap mechanism to video editing, extending its capabilities beyond static images. Both Null-text Inversion \cite{Mokady_2023_CVPR} and BLIP-Diffusion \cite{li2024blip} completely follow the operational framework established by P2P. In Null-text Inversion, the generation process is guided by both a source prompt and an edit prompt, while BLIP-Diffusion relies on a combination of a subject image and edit text. StyleDiffusion \cite{li2023stylediffusion} introduces P2Plus, which modifies not only the self-attention maps of the conditional branch in diffusion models but also those of the unconditional branch. P2P and its derivatives are effective in spatial control, enabling precise adjustments with minimal computational cost. However, these methods struggle when editing multiple elements that need to be seamlessly integrated. Additionally, they are primarily focused on text-based editing, their effectiveness in handling multi-modal inputs for more complex multimodal tasks has yet to be explored. Unlike P2P, a new method called FAM Diffusion\cite{yang2024fam} focuses on the challenge of high-resolution image generation and innovatively proposes an attention map modulation module. This module performs a weighted fusion of low-resolution and high-resolution attention maps to control local texture details, thus enabling the generation of high-quality and high-resolution images. However, when dealing with spatial details in complex scenes, this method may fail to precisely capture and generate the fine spatial details of various regions, resulting in blurred or inaccurate local textures.

\subsubsection{Selective Local Attention Composition}
\label{sec:Selective Local Attention Composition}
Rather than manipulating full attention maps, selective local attention composition \cite{lu2023tf,patashnik2023localizing} selectively integrates portions of the attention map into a newly synthesized map, focusing on specific patches or pixels of the image. This method is shown in Fig.~\ref{fig:class2}(d). This method is dedicated to preserving locally desired features from cross-attention and self-attention, which is beneficial to cross-domain image synthesis. TF-ICON \cite{lu2023tf}, designed for training-free cross-domain image-guided composition, introduced a self-attention composition method. The composite self-attention map consists of three parts: self-attention from the reference and background images, along with a cross-attention map calculated between them based on patch indices. To refine specific shapes, Object-Shape Variations \cite{patashnik2023localizing} selectively fuses the rows and columns of the source image’s self-attention map that correspond to the pixels containing the object of interest into the self-attention map of the generated image, utilizing a mask guidance mechanism. Selective local attention composition methods offer strong spatial control by focusing on specific regions or patches of an image. This allows for fine-tuned modifications that preserve foreground details and adapt to different domains. However, their reliance on local adjustments can limit global spatial coherence. This may result in unnatural transitions or inconsistencies, especially in complex scenes.

\begin{figure*}[ht]
\centering
\includegraphics[width=1.0\textwidth]{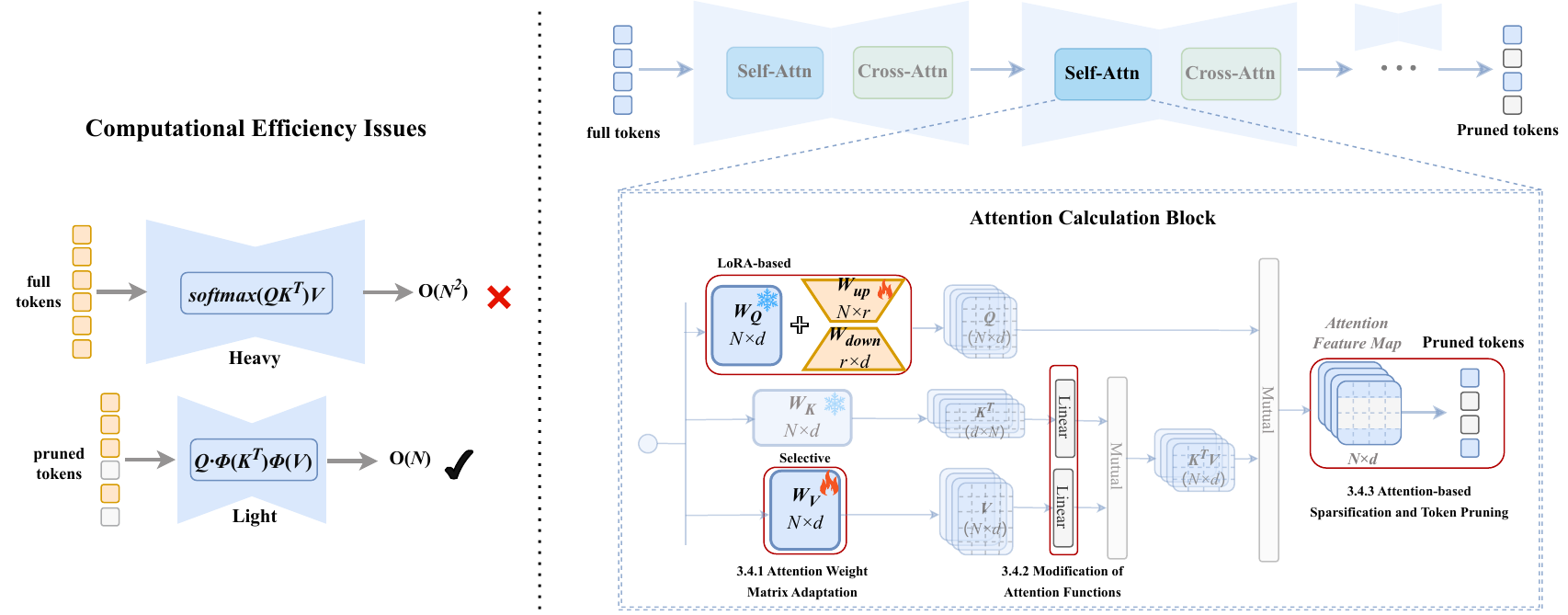}
\caption{\label{fig:class4}Efficiency issues in diffusion models are shown on the left and a common pipeline of attention-based modification of computational efficiency is illustrated on the right. The modulated components of attention are highlighted with red boxes.}
\end{figure*}

\subsection{Temporal Fusion}
\label{sec:Temporal Fusion}
Temporal features, which capture implicit movement information in time-dependent data such as videos, are crucial for ensuring temporal consistency during the generation process. To fully leverage these temporal features in diffusion models, two primary approaches have been proposed: Temporal attention injection and spatio-temporal feature alignment. These methods help integrate and align temporal information effectively to ensure high-quality generation in tasks like video generation, which is shown in Fig.~\ref{fig:class3}

\subsubsection{Temporal Attention Injection}
\label{sec:Temporal Attention Injection}
In this method, a dedicated temporal attention layer is directly inserted into the diffusion model's architecture to capture the temporal dependencies in sequential data. This method often works in conjunction with spatial attention, where temporal attention layers are introduced to understand movement dynamics across frames while preserving spatial coherence.

For example, factorized spatio-temporal attention layers stack a temporal attention layer following a spatial attention layer, enabling the model to dynamically adjust to time-related features. This approach has been implemented in models such as VDM \cite{ho2022video}, Structure and Content-Guided Video\cite{Esser_2023_ICCV}, Imagen \cite{ho2022imagen}, and Make-a-Video \cite{singer2022make}. These models apply temporal attention to determine when and how to focus attention across time sequences, ensuring that the temporal relationships are effectively captured.

\subsubsection{Spatio-Temporal Feature Alignment}
\label{sec:Spatio-Temporal Feature Alignment}
The spatio-temporal feature alignment approach emphasizes computing cross-attention between different frames to align temporal features more effectively. This method replaces traditional self-attention maps with cross-attention, establishing correspondences between the previous and current frames to guide the generation process. Models like Pix2Video\cite{ceylan2023pix2video}, Tune-A-Video \cite{wu2023tune}, Video-P2P \cite{liu2024video}, VideoComposer\cite{NEURIPS2023_180f6184} and Text2Video-Zero \cite{Khachatryan_2023_ICCV} use cross-attention mechanisms between frames to improve temporal consistency and alignment in video generation. In particular, VideoGrain \cite{yang2025videograin} advances this direction by modulating both spatio-temporal cross-attention and self-attention to achieve fine-grained text-to-region control and feature separation. Cross-attention is refined to localize each prompt to its corresponding region, while self-attention is adjusted to preserve intra-region consistency and suppress inter-region interference.

\subsection{Computational Efficiency}
\label{sec:Computational Efficiency}
As diffusion models evolve to handle more complex multimodal data, the computational demands on attention mechanisms have become increasingly significant. Addressing this challenge, recent advancements focus on optimizing attention mechanisms to balance performance and computational efficiency. The following methods have been proposed to enhance computational efficiency in diffusion models: attention weight matrix adaptation, modification of attention functions, and sparsification and token pruning. The common pipeline of these methods are shown in in Fig.~\ref{fig:class4}.

\subsubsection{Attention Weight Matrix Adaptation}
The approach outlined in this section involves fine-tuning the weight matrix of the attention layer through training, which enables the model to learn quickly with a small amount of data and improve the performance based on the original pre-trained model. There are two common approaches to fine-tuning in Fig.~\ref{fig:class4}. One is to introduce a new adapter like Low-rank adaptation (LoRA), and the other is to select partially existing parameters.
\label{sec:Attention Weight Matrix Adaption}
\begin{itemize}
    \item \textbf{LoRA-based Finetuning}
\end{itemize}

LoRA \cite{hu2021lora,he2023efficientdm,dettmers2024qlora,smith2024continual,liang2024inflora,liu2024dora,zhangparameter,chen2023longlora,yang2024lora} widely regarded as a parameter-efficient fine-tuning method, introduces trainable low-rank decomposition matrices into a diffusion model while keeping the pre-trained model weights frozen. In principle, LoRA can be applied to any subset of a neural network’s weight matrix, significantly reducing the number of trainable parameters required for adaptation. In classical LoRA \cite{hu2021lora}, the method is used to adapt the weight matrix of self-attention layers within the Transformer architecture during experiments. By significantly reducing the number of trainable parameters required for specific tasks, LoRA makes training more efficient. As a result, many LoRA-based variants \cite{dettmers2024qlora,smith2024continual,liang2024inflora,liu2024dora,zhangparameter,chen2023longlora,yang2024lora} have gained popularity in the fine-tuning research of diffusion models. Some of these approaches \cite{dettmers2024qlora,liang2024inflora,liu2024dora,zhangparameter,chen2023longlora} apply LoRA to adapt the self-attention layers, while others \cite{smith2024continual,yang2024lora} focus on the cross-attention layers. Notably, AnimateDiff \cite{guo2023animatediff} inserts trainable weight matrices into both self-attention and cross-attention layers.

\begin{itemize}
    \item \textbf{Selective Finetuning}
\end{itemize}

Instead of training the entire attention layer or inserting additional networks, the selective fine-tuning method in Fig.~\ref{fig:class4} targets the cross-attention layer and selectively fine-tunes specific parameters while freezing most of them. Typical examples are custom diffusion \cite{kumari2023multi} and Continual Diffusion\cite{smith2024continual}, which freeze the $W_q$ matrix and fine-tune $W_k$ and $W_v$ to to reduce the number of parameters to be trained.

\subsubsection{Modification of Attention Functions}
While attention-based models are renowned for their excellent parallel performance, they inherently face both space and time complexity challenges of $O(n^2)$. As the sequence length $n$ increases, the computational demands of the attention layer rise significantly. Recently, several approaches, as is illustrated in Fig.~\ref{fig:class4}, have been proposed to reduce the computational complexity of attention by modifying its structure and the underlying formula. Although many of these methods were not initially designed for diffusion models, an increasing number of studies have successfully adapted them for diffusion model applications. In this subsection, the improvement of computational efficiency on the software and hardware level will be presented separately.

\begin{itemize}
    \item \textbf{Linear Attention Computation}
\end{itemize}

The softmax attention mechanism, introduced by the Transformer model \cite{vaswani2017attention}, has seen significant development in recent years due to its high performance. However, the computational complexity of softmax attention is $O(n^2)$, and directly calculating self-attention often results in high computational costs. To address this issue, linear attention \cite{katharopoulos2020transformers,lu2021soft,shen2021efficient,han2023flatten,qin2022cosformer,choromanski2020rethinking} has been proposed. Unlike softmax attention, linear attention decouples the softmax function into two independent functions, allowing the order of computation to be adjusted from $Softmax(Q \cdot K) \cdot V$ to $Q \cdot (\phi(K) \cdot \phi(V))$ or $(\phi(Q) \cdot \phi(K)) \cdot V$, where $\phi(\cdot)$ represents a linear function. While linear attention was originally developed for Transformers, the rise of diffusion models has led to increasing research on applying linear attention to SD \cite{zhu2024dig,han2024agent,xie2024sana}. Notably, Agent Attention \cite{han2024agent} combines the advantages of both softmax and linear attention, further enhancing performance.

\begin{itemize}
    \item \textbf{Hardware-based Chunk Attention}
\end{itemize}
In diffusion models, the attention mechanism often faces challenges related to high memory access costs and low throughput. Hardware-software co-design techniques optimize the utilization of hardware resources and improve algorithm efficiency, significantly enhancing computational performance. This combination allows for more effective processing of large-scale data, especially in complex tasks like those in diffusion models.
In the standard attention mechanism, data is transferred from the slower High Bandwidth Memory (HBM) to Static Random Access Memory (SRAM) for processing, then returned to HBM after computation. Accessing HBM for computation incurs significant costs. To address this issue, Flash Attention \cite{dao2022flashattention} divides the input matrices $Q$, $K$, and $V$ into smaller blocks, loading these blocks from GPU memory (HBM) into fast cache (SRAM) and performing attention operations on each block before updating the output in HBM. This method, known as tiling, reduces memory read and write operations, leading to computational acceleration. However, despite these improvements, overall throughput remains low. Flash attention v2 \cite{dao2023flashattention} was introduced to further enhance throughput by building on the advancements of Flash Attention. Flash attention v2 optimizes the chunking strategy by assigning each thread block the responsibility of computing one attention head for a specific block. Within each thread block, multiple warps of threads work together to perform matrix multiplication. Unlike Flash attention v1, which employed a general chunking approach, Flash attention v2 focuses its chunking strategy on $Q$. This method allows the final result to be obtained by concatenating the outputs of each block, eliminating the need for inter-warp communication and reducing additional operations along with the associated read and write processes. Consequently, the chunking strategy in Flash Attention v2 is more efficient. Similarly, GLA \cite{yang2023gated} combines linear attention with selective forgetting and chunk-wise block-parallel attention, enabling efficient parallel training on tensor cores. It doesn’t specifically target diffusion models but can be generalized to Denoising Diffusion Transformers. In conclusion, flash attention focuses on optimizing memory access patterns, while GLA leverages efficient parallel computation. Flash Attention v2, with its refined chunking strategy, provides a significant improvement in throughput, but may still face challenges in extreme-scale applications. On the other hand, GLA’s parallel processing capabilities have scalability for large-scale training, but selective forgetting could limit its performance in certain scenarios. Both methods represent significant advancements in the pursuit of faster, more efficient attention computations, but their effectiveness depends on the specific use case and requirements. FlashMask\cite{wang2024flashmask} builds upon Flash Attention by introducing a column-wise sparse mask representation, which enables optimized kernel implementations to efficiently detect and bypass redundant computations within masked regions. This design specifically targets the limitations of Flash Attention in processing complex or structured attention masks. By leveraging sparsity at the column level, FlashMask reduces the memory complexity to linear.

\subsubsection{Attention-based Sparsification and Token Pruning}
\label{sec:Attention-based Sparsification and Token Pruning}

Model compression can be divided into two categories: a) attention sparsification\cite{yuan2024ditfastattn,su2024f3} and b) token pruning\cite{wang2024zero,wang2024attention,smith2023coda,Bolya_2023_CVPR} based on the attention map. Specifically, our paper only discusses compression methods related to attention layers. Rather than compressing each each parameter in the weight matrices, sparsification and pruning operate on the principle of eliminating unimportant weights, thereby reducing the number of parameters and computational load while maintaining accuracy. Attention sparsification focuses on reducing the parameters of the attention map, while token pruning involves removing input tokens that contribute little to the prediction. DiTfastattn\cite{yuan2024ditfastattn} is a typical method of attention map sparsification. DiTFastAttn achieves attention sparsification through three techniques. First, window attention with residual sharing reduces spatial redundancy by applying window attention and caching residuals. Second, attention sharing across timesteps skips computations by leveraging the similarity of attention outputs between adjacent timesteps. Third, the reuse of attention outputs during unconditional generation, based on the similarity between conditional and unconditional inferences, avoids redundant computations. Moreover, F$^3$-pruning \cite{su2024f3} builds upon the temporal attention used in models like CogVideo and Tune-A-Video, introducing a pruning strategy to remove redundant temporal attention in later stages of video generation. The pruning process identifies temporal attention weights with lower aggregate attention scores, which are considered less important and are pruned, optimizing the model's efficiency. As typical examples of token pruning, Zero-TPrune\cite{wang2024zero} and AT-EDM\cite{wang2024attention} use a graph-based pruning layer placed after the attention layers. This layer treats the attention matrix as an adjacency matrix of a complete directed graph, with tokens as nodes and attention as edges, to obtain an importance score distribution on tokens and retain the top-k important tokens. Similarly, CODA-Prompt\cite{smith2023coda} introduces a novel attention-based prompt selection method, which generates prompts passed through multiple layers of a large-scale pre-trained model. Specifically, instead of removing parts of tokens, ToMe\cite{Bolya_2023_CVPR} introduces a merging mechanism to reduce the number of tokens, inserting a merge layer before each self-attention and cross-attention layer. VidToMe\cite{li2024vidtome} extends the token merging mechanism to video generation by integrating merged tokens prior to self-attention layers and performing subsequent unmerging operations. This architectural innovation enhances computational efficiency while facilitating spatio-temporal consistency. In 3D scene editing, EditSplat\cite{in2024editsplat} assigns attention weights to each Gaussian by back-projecting the cross-attention maps between text and image onto 3D Gaussians. Redundant Gaussians are pruned and selectively optimized based on these weights, enabling efficient optimization and semantically localized editing, thereby enhancing 3D editing performance.

\section{Related Applications}\label{sec:Related Applications}
In this section, we will explore various applications of attention in diffusion models, ranging from unimodal to multimodal tasks. Attention mechanisms have been increasingly integrated into these models to enhance their performance in diverse areas. Some methods leverage the inherent attention mechanisms of the diffusion model, while others modify these mechanisms. For methods that involve modifications, which were introduced in Section~\ref{sec:Roles and Modulation Methodologies of Attention in Diffusion Models}, further details will not be repeated. However, for methods that solely utilize the inherent attention, not discussed in Section~\ref{sec:Roles and Modulation Methodologies of Attention in Diffusion Models}, a detailed description will follow. These methods show both the potential benefits and the challenges involved. By focusing on the integration of attention and diffusion processes, this section will provide new insights and solutions for practical applications.

\subsection{Unimodal Learning}
\subsubsection{Image Translation and Inpaiting}
In image-to-image translation tasks, traditional methods often require customized hyperparameters or network structures for each specific task, lacking a unified approach. Palette\cite{saharia2022palette} provides a unified framework that eliminates the need for task-specific adjustments. It leverages conditional diffusion models integrated with self-attention mechanisms to handle a variety of image translation tasks, including colorization, inpainting, uncropping, and JPEG restoration. Additionally, Palette introduces a unified evaluation protocol based on ImageNet and Places2 to consistently assess these diverse tasks. Similarly, SEMIT\cite{wang2020semi} proposes an image-to-image translation method based on semi-supervised learning and few-shot learning. By combining a small amount of labeled data with a large amount of unlabeled data, along with pseudo-label generation and cycle consistency constraints, SEMIT achieves high-quality image translation without the need for extensive labeled datasets. Furthermore, Pix2Pix-Zero \cite{parmar2023zero} proposes a zero shot image-to-image translation method called Pix2Pix-Zero, which eliminates the need for manual text prompts or additional training. Using a cross-attention guidance mechanism, this approach preserves the structure of the input image during the diffusion process, maintaining the layout and object consistency while transforming the content to align with the target domain.

For image inpainting specifically, traditional methods often require training on specific mask distributions, making it challenging to generalize to free-form or extreme mask scenarios (e.g., large missing areas). Furthermore, GANs and other generative methods frequently produce simple textures lacking semantic coherence for large-scale inpainting tasks, and the boundaries between the generated and known regions are often inconsistent or discontinuous. To address these limitations, \cite{lugmayr2022repaint} introduces an image inpainting method based on DDPM. By leveraging conditional constraints during the reverse diffusion process, it progressively transforms random noise into inpainting results that align with the original image distribution. This approach eliminates the need for retraining on task-specific data, enabling high-quality and diverse free-form image inpainting.

\subsubsection{Image Super-resolution}
Traditional diffusion models perform well in generating low-resolution images but still exhibit significant performance gaps in high-resolution generation tasks. \cite{ho2022cascaded} and \cite{saharia2022image} use cascade architectures to progressively enhance image resolution. \cite{ho2022cascaded} introduces cascaded diffusion models, which employ a cascaded structure and incorporate conditioning augmentation to inject noise into the input data, simulating distribution shifts. This approach prevents error accumulation during the cascaded generation process, enabling the production of higher quality, high-resolution images. However, it does not explicitly modify or optimize the attention mechanisms, despite using multi-head self-attention layers. The general self-attention layers in the model have limits to the unique challenges of high-resolution image generation, such as capturing fine details and handling large-scale spatial dependencies. On the other hand, \cite{saharia2022image} relies on low-resolution images as conditional inputs, with each stage of generation being strictly constrained by the low-resolution input. This method directly extracts information from the low-resolution image, placing greater emphasis on pixel-level consistency with the input. However, the lack of proper attention mechanism limits the model's capacity to adaptively prioritize relevant image features at different scales, potentially restricting its ability to refine high-frequency details.

\subsubsection{Style Transfer}
Style transfer involves blending the content of one image with the style of another to create a new image that preserves the original content while adopting the new style. This process typically consists of two main steps: preparing the content and style images, and generating the new image by extracting features through a diffusion model and optimizing loss functions. In these models, attention mechanisms could play a crucial role in selectively focusing on the relevant parts of the content and style images. Z-STAR \cite{deng2023z} focuses on improving the fusion of content and style in the latent space by leveraging cross-attention feature rearrangement within diffusion models. During the denoising process, cross-attention aligns features from the content image with those from the style image, guiding the diffusion process to effectively combine style and content. Zecon \cite{yang2023zero} introduced a patch-wise contrastive loss, guided by attention mechanisms, to focus on individual patches of the image. This loss computes similarities between patches of the content image and the generated image, maximizing mutual information in regions where content needs to be preserved. Additionally, the attention mechanism is enhanced by a directional loss in the CLIP model, which aligns the text description of the style with the content features.

\subsubsection{Detection}
In the field of computer vision, as task complexity and application demands continue to grow, higher standards are being set for detection tasks across various scenarios. Here, we focus on four types of detection tasks—object detection, out-of-distribution (OOD) detection, temporal action detection, and diffusion-generated image detection—and introduce four corresponding studies. DiffusionDet\cite{chen2023diffusiondet} redefines object detection as a denoising process from noisy boxes to target boxes, breaking the reliance on fixed prior frameworks in traditional detection methods and significantly improving adaptability and performance in sparse or crowded scenarios. DIFFGUARD\cite{gao2023diffguard} leverages the conditional generation capabilities of diffusion models to amplify the semantic differences between the input image and the conditionally generated image, achieving effective OOD detection, particularly excelling on large-scale datasets like ImageNet. \cite{wang2023dire} measures the error between the input image and its reconstruction by a pre-trained diffusion model, utilizing the difference in reconstruction errors between real and diffusion-generated images to provide a powerful tool for detecting diffusion-generated images, with exceptional performance even on samples from unseen diffusion models. Most existing detection algorithms have benefited from the integration of diffusion models. Unfortunately, few of them explored the role of attention mechanisms in detection. DiffTAD\cite{nag2023difftad} introduces an attention-based framework for temporal action detection using proposal denoising diffusion. It progressively generates action boundaries to resolve temporal ambiguity, improving detection accuracy and efficiency. Attention mechanisms help capture key temporal features by focusing on relevant time segments, enhancing the model's ability to track actions accurately. In DiffTAD, the model selects a subset of queries based on pairwise similarity and IoU measurement in an attention-based manner. This approach extends the application of attention to complex temporal detection tasks, enabling more accurate and efficient action detection over time. Therefore, the application of attention in diffusion models for detection tasks is an area that warrants further exploration by researchers in the future.

\subsubsection{Unimodal Image Segmentation}

Semantic segmentation aims to classify every pixel in an image, assigning each pixel a semantic category label to generate a pixel-level segmentation map. However, as a dense pixel-level prediction task, semantic segmentation requires pixel-wise annotations, which are not only time-consuming and labor-intensive but also prone to errors. Additionally, most current mainstream methods rely on fully supervised pretraining, which performs well on large annotated datasets (e.g., ImageNet classification datasets) but struggles in low-annotation scenarios. Attention mechanisms in diffusion models could help address this challenge by allowing the model to focus on important regions of the image, improving segmentation performance with fewer annotations. By leveraging attention, the model could dynamically prioritize pixel-level features, enhancing its ability to handle low-annotation tasks more efficiently. Existing methods use diffusion models as tools. They rely on the inherent attention mechanism to aid segmentation but make little to modifications. \cite{asiedu2022decoder} introduces the Decoder Denoising Pretraining (DDeP) method, which compensates for the limitations of randomly initialized decoders. By combining a supervised pretrained encoder with a denoising pretrained decoder, DDeP enables efficient end-to-end fine-tuning. For the issues of high computational cost and slow inference speed in traditional diffusion models, \cite{ji2023ddp} proposes a general framework (DDP) based on conditional diffusion models. This framework improves model efficiency for tasks such as semantic segmentation, depth estimation, and BEV map segmentation through a decoupled design and a lightweight map decoder module. Furthermore, to reduce the dependency on external pretrained models and improve performance on small datasets, \cite{amit2021segdiff} introduces a segmentation framework based on conditional diffusion probabilistic models. By integrating image features with segmentation estimation features during the stepwise denoising generation process, SegDiff employs a lightweight encoder-decoder structure (U-Net) to generate high-quality segmentation masks. It also uses a multiple generation strategy to enhance the stability and accuracy of results, achieving improved performance on small datasets and in multi-domain tasks, such as urban scenes, medical images, and remote sensing images.

\subsubsection{Image classsification}
The goal of image classification is to assign one or more category labels to an entire image based on its content, providing critical support for other tasks such as object detection and image segmentation. \cite{mukhopadhyay2023diffusion} repositions diffusion models for classification tasks, analyzing in detail how to extract features from different stages of the diffusion process to optimize classification performance. On several fine-grained classification datasets (e.g., Aircraft, CUB, Flowers), diffusion model features demonstrate strong transferability. However, attention mechanisms have not been specifically optimized to enhance the model’s ability to focus on key image regions, which could improve classification accuracy, especially in complex tasks. \cite{chen2023robust} leverages diffusion models to build a robust diffusion classifier, enhancing the model's defense against adversarial examples and improving its generalization ability. To further apply diffusion models to classification tasks and even zero-shot learning, \cite{li2023your} proposes a novel approach that combines the density estimation capability of generative models with classification tasks, achieving impressive results in scenarios such as zero-shot learning, multimodal reasoning, and out-of-distribution generalization. In this method, cross-attention is used for semantic alignment between text and images. Integrating attention in the diffusion classifier could further enhance the model's focus on critical features, improving its adaptability and performance in challenging classification tasks.

\subsection{Multimodal learning}
\subsubsection{Text-to-Image Controllable Generation}
Text-to-Image controllable generation refers to the task of generating images with specific attributes and details based on textual descriptions. The primary objective of this task is to ensure that the generated images not only align with the content of the text but also maintain high visual quality and consistency. Text-to-Image controllable generation has two key challenges: consistency enhancement and spatial control. Consistency enhancement ensures that the generated image stays faithful to the text, preserving coherence in attributes such as color, object identity, and their relationships. Meanwhile, spatial control adjusts the positioning and arrangement of objects within the image, ensuring their placement aligns with the text's description.
To address these challenges, attention mechanisms in diffusion models are commonly employed. Several methods in this area focus on modifying attention at different levels. At the attention feature level, techniques like self-attention feature injection, conditional alignment in cross-attention, and selective local attention composition intervene with the input textual and visual features at the attention layer. These modifications, including MasaCtrl\cite{cao2023masactrl}, DreamMatcher\cite{nam2024dreammatcher}, PnP\cite{tumanyan2023plug}, Fec\cite{10424833}, eDiff-I\cite{balaji2022ediff}, IP-Adapter\cite{ye2023ip}, and InstanceDiffusion\cite{wang2024instancediffusion}, ensure that the generated image meets the desired attributes as specified by the text.
In contrast, attention map level modulation methods, such as P2P\cite{hertz2022prompt}, Null-text Inversion\cite{mokady2023null}, StyleDiffusion\cite{li2023stylediffusion}, BLIP-Diffusion\cite{li2024blip}, Object-Shape Variations\cite{patashnik2023localizing}, and TF-ICON\cite{lu2023tf}, adjust the full or partial cross-attention maps to enhance the alignment between the text and the generated image. Additionally, methods like BoxDiff\cite{xie2023boxdiff}, CDS\cite{nam2024contrastive}, Predicated Diffusion\cite{sueyoshi2024predicated}, Energy-Based Cross Attention\cite{park2024energy}, FoI\cite{guo2024focus}, and Shape-Guided Diffusion\cite{park2024shape} focus on using attention maps to impose additional constraints, further refining the generation process to ensure that the output not only aligns with the text but also adheres to specific constraints and conditions.

\subsubsection{Multimodal Image Segmentation}
Multimodal image segmentation involves segmenting an image by incorporating information from multiple modalities \cite{chen2022semi,chen2024dynamic,zhang2021siamcda,zhang2024amnet}. The goal is to utilize complementary features from each modality to improve the accuracy and robustness of the segmentation process, thereby offering a more comprehensive understanding of the image's content. Diffusion models with attention, originally designed for image generation, can be adapted to this task by refining multimodal inputs during the denoising process. By applying attention mechanisms, these models focus on the most relevant features from each modality, improving the integration of spatial and contextual information and enhancing segmentation accuracy. Some methods\cite{pnvr2023ld,zhao2023unleashing} utilize the inherent attention layers of LDM for segmentation, while others \cite{wu2023diffumask} usually adopt attention-based mask guidance. For instance, LD-ZNet \cite{pnvr2023ld} maps the segmentation task to the latent space of the diffusion model, aligning intermediate semantic features with the provided text prompts. It incorporates a lightweight ZNet and an enhanced LD-ZNet module, which effectively fuse latent features using cross-modal attention, improving segmentation performance for both real-world and AI-generated images. Similarly, VPD \cite{zhao2023unleashing} explores how pretrained text-to-image diffusion models can transfer multimodal semantic knowledge to tasks like semantic segmentation. It utilizes denoising networks and cross-attention mechanisms to extract visual features and semantic alignment, enhancing segmentation accuracy with lightweight text adapters and task-specific decoders. Meanwhile, to address challenges like high annotation costs and limited generalization, DiffuMask \cite{wu2023diffumask} generates high-quality pixel-level semantic masks by utilizing cross-attention maps from the diffusion model. It further refines these outputs using multi-resolution fusion, adaptive thresholding, dense conditional random fields, and data augmentation, reducing annotation costs and enhancing segmentation performance.

\subsubsection{Text-to-Video Generation}
Text-to-Video (T2V) aims to generate entire video sequences that align with the content, context, and motion described in the text. While the text input typically describes static scenes or events, video generation requires converting these descriptions into dynamic processes. This necessitates the use of attention mechanisms in diffusion models to simultaneously handle spatial information (the details of each frame) and temporal information (the coherence between frames). In this field, methods like temporal attention injection and spatio-temporal feature alignment are commonly employed at the attention feature level. These techniques are used by approaches such as VDM\cite{ho2022video}, Text2Video-Zero\cite{Khachatryan_2023_ICCV}, Make-A-Video\cite{singer2022make}, VideoComposer\cite{NEURIPS2023_180f6184} and Imagen\cite{ho2022imagen} to enhance the alignment of both spatial and temporal features, ensuring a smooth and contextually consistent video generation process.

\subsubsection{Video Editing}
Video editing involves the precise modification and replacement of objects, scenes, or specific regions within a video by leveraging text prompts, target images, and other conditions, while maintaining temporal and visual consistency across frames. Various works have proposed innovative techniques to address these challenges. RAVE \cite{kara2024rave} introduces a noise shuffling strategy that enhances spatio-temporal interactions between video frames, enabling efficient zero-shot editing with a pre-trained text-to-image diffusion model, significantly improving editing speed while ensuring temporal consistency, even for long and complex videos. Similarly, Pix2Video \cite{ceylan2023pix2video} builds on a depth-conditioned image diffusion model and employs self-attention feature injection along with guided latent variable updates to achieve text-driven video editing with consistent appearance and geometry across frames. StableVideo \cite{chai2023stablevideo} further improves temporal consistency in video editing by introducing an inter-frame propagation mechanism and layered representations, which ensure stable and geometry-consistent object editing with smooth transitions and high fidelity. Extending beyond individual tasks, VIDiff \cite{xing2023vidiff} presents a unified multi-modal diffusion framework to tackle multi-task support, long video editing, and inference efficiency. It incorporates a multi-modal condition injection mechanism for text and image inputs, temporal attention layers to enhance cross-frame consistency, and an iterative inference approach to enable efficient and consistent editing for long videos.

Building on traditional video editing techniques, some methods have shifted their focus towards fine-grained and precise editing by leveraging text prompts and target image information to enhance the control and quality of edits. GenVideo \cite{harsha2024genvideo} employs shape-aware mask generation and latent noise correction strategies to achieve accurate object editing within videos. By maintaining temporal consistency across frames, it delivers high-quality results even for challenging scene modifications, showcasing its robustness in detailed and complex video editing tasks.

Traditional video editing methods often rely on extensive labeled data and task-specific training, which can be time-consuming and resource-intensive. In contrast, zero-shot video editing provides a flexible and efficient solution by eliminating the need for such resources. FateZero \cite{qi2023fatezero} introduces a novel attention fusion mechanism to capture motion and structure information during the reverse diffusion process, enabling zero-shot text-driven editing of attributes, style, and shape with temporal consistency and high-quality results across frames. VidToMe \cite{li2024vidtome} further advances zero-shot editing by focusing on improving temporal consistency through the fusion and compression of cross-frame self-attention tokens. This strategy not only reduces computational complexity but also ensures high-quality frame generation in text-driven video editing. In contrast, CAMEL \cite{zhang2024camel} takes a parameter-efficient fine-tuning approach by introducing causal motion-enhanced attention mechanisms and learnable motion prompts, which require optimization specific to the input video. By disentangling and refining motion dynamics and appearance content, it achieves improved motion coherence and maintains consistency across a wide range of editing scenarios, making it a highly flexible yet not strictly zero-shot approach.

\subsubsection{3D Reconstruction}
3D reconstruction\cite{wang2024gaussianeditor,liu2024dynvideo,yang2024dreamcomposer} is the task of generating a model that accurately reflects the true three-dimensional geometry of an object or scene by extracting depth and structural information from one or more 2D images. \cite{wang2024gaussianeditor} addresses the challenges of precise localization and control in 3D scene editing using existing 2D diffusion models, it introduces a systematic framework based on 3D Gaussian distributions, enabling fine-grained editing of 3D scenes through text instructions, significantly improving the precision and effectiveness of editing while reducing training time. Additionally, to overcome the lack of consistency in traditional 2D representations when dealing with large-scale motion and view changes, \cite{liu2024dynvideo} proposes a video editing framework based on dynamic NeRF, this framework integrates 2D and 3D diffusion priors to achieve highly consistent and finely detailed editing of videos featuring large-scale motion and view changes.

\subsubsection{3D Editing}
3D editing refers to the process of modifying, adjusting, and optimizing existing three-dimensional models to achieve specific visual effects or functional requirements. \cite{pandey2024diffusion} introduces a novel method called "Diffusion Handles," which lifts the activations of diffusion models into 3D space to enable fine-grained, 3D-aware editing of objects in 2D images, without requiring additional training or 3D data. GaussianEditor\cite{chen2024gaussianeditor} presents a 3D editing algorithm named GaussianEditor, which leverages semantic tracing and hierarchical Gaussian splatting to achieve efficient and detailed editing and repair of 3D scenes within a short time. \cite{yenphraphai2024image} offers a new approach to image editing by combining 3D geometry control with the generative capabilities of diffusion models, providing a complete process from coarse deformation to high-fidelity image generation, thereby enhancing precision and flexibility in the field of image editing. Lastly, EditSplat\cite{in2024editsplat} proposes the Multi-View Fusion Guidance (MFG) and Attention-Guided Trimming (AGT) methods. MFG projects and fuses multi-view images using the depth maps of 3DGS and ensures that the editing is consistent with multi-view information by leveraging classifier-free guidance. AGT assigns weights to 3D Gaussians based on the attention maps of the diffusion model. It prunes Gaussians with high weights and selectively optimizes them, thus improving optimization efficiency and semantic local editing capabilities.

\subsection{Other tasks}
The emergence and evolution of recommendation tasks are intrinsically tied to the rapid advancements in information technology and the internet. These tasks are extensively utilized in domains such as e-commerce, social media, music and video streaming, and online education. By analyzing users' behaviors, preferences, and contexts, recommendation systems strive to identify and deliver the most relevant content or items from a vast pool of information to fulfill users' needs. In the context of single-modality recommendation tasks, real-world social relationships often contain a significant amount of irrelevant or false social links, known as noise, which can corrupt user embeddings and degrade recommendation performance. To tackle this challenge, RecDiff \cite{li2024recdiff} introduces a social recommendation framework based on diffusion models. Its core mechanism lies in multi-step diffusion and denoising within the latent space, which improves the accuracy of user preference representations and enhances recommendation performance. On the other hand, for multimodal recommendation tasks, where leveraging item information from multiple modalities is key to overcoming data sparsity and boosting recommendation accuracy, MCDRec \cite{ma2024multimodal} proposes a multimodal conditioned diffusion model. This framework utilizes the generative capabilities of diffusion models to seamlessly integrate multimodal information (e.g., visual and textual features) with user collaborative signals, while simultaneously denoising the user behavior graph. Despite their different focuses—single-modality for RecDiff and multimodality for MCDRec—both methods effectively harness diffusion models to address key challenges in recommendation tasks.

\section{Challenges and Future Directions}\label{sec:Challenges and Directions}
Despite the success achieved in attention mechanism with diffusion models, there are still challenges that need to be addressed in future work.

\subsection{Diffusion Models for Discriminative Tasks}
Diffusion models have demonstrated exceptional performance in generative tasks, excelling in the creation of high-quality images, text, and other forms of content. However, applying these models to discriminative tasks requires strong recognition and classification capabilities, which remains a significant challenge. Extracting meaningful features and achieving precise classification within this context highlights the limitations of diffusion models when directly applied to discriminative objectives.

Discriminative tasks typically require explicit labels and supervised learning signals, whereas diffusion models are predominantly trained for generative purposes using unsupervised or self-supervised strategies. This divergence raises an important question: how can diffusion models be effectively adapted to leverage supervised signals and achieve competitive performance in discriminative tasks? Addressing this requires innovations in both model architecture and training methodologies to bridge the gap between generative and discriminative paradigms.

Despite these challenges, the proven success of diffusion models in multimodal generative tasks underscores their vast potential. Advancements in computational efficiency, enhanced multimodal learning techniques, and innovative training strategies pave the way for applying diffusion models to discriminative tasks. Continued research is anticipated to unlock their full potential, positioning diffusion models as a transformative tool for cross-modal and discriminative applications.

\subsection{Semantic Consistency}
The feature injection methods, including both self-attention and cross-attention feature injection discussed in Section \ref{sec:Consistency Enhancement}, have demonstrated impressive performance across a wide range of editing tasks, such as object replacement, addition, removal, action editing, scene editing, style editing, and more. Notably, these methods excel at maintaining the consistency between the edited and original images. However, since different editing tasks prioritize different types of consistency, the effectiveness of these methods is often task-specific. For instance, some methods focus primarily on spatial layout consistency, which limits their ability to perform tasks like action editing. In contrast, Kv Inversion\cite{huang2023kv} and MasaCtrl\cite{cao2023masactrl} consider texture and identity consistency, enabling more complex edits. Unfortunately, Kv Inversion and MasaCtrl struggle when there are significant incompatibilities between prompts and images or when the layout changes dramatically. Z-STAR\cite{deng2023z} focuses exclusively on style editing, while PnP\cite{tumanyan2023plug} encounters difficulties when editing small images without texture. Future work should focus on improving semantic consistency across diverse tasks to broaden the applicability of these methods.

\subsection{Precise Controllable Editing}
Among the methods discussed in Section~\ref{sec:Spatial Control}, cross-attention map control has emerged as the most effective pipeline for detailed image editing. Following P2P\cite{hertz2022prompt}, which pioneered controllable editing, numerous studies have built upon this approach. However, these methods face common challenges. First, cross-attention map control strategies, like P2P, require exact alignment between the source prompt and the target prompt, which imposes significant limitations and hinders editing efficiency. Second, the generation process fails if the target prompt includes unknown content or unseen object parts in the source image. The accurate localization of text embeddings through cross-attention mapping to the visual space remains a major challenge. Therefore, an important future research direction is to enable precise and efficient control over editing content through cross-attention maps, even in scenarios where objects in the image are unknown or partially invisible.

\subsection{Computation Acceleration}
The standard attention mechanism suffers from high time complexity and low computational efficiency due to the computation of the Softmax function. At the software level, while the incorporation of linear attention\cite{han2024agent}, as discussed in Section~\ref{sec:Computational Efficiency}, significantly reduces computational complexity and enhances the model's efficiency for handling long sequences, it also introduces performance degradation and additional computational overhead, which partially offsets the efficiency gains. On the hardware level, chunk attention\cite{dao2022flashattention,dao2023flashattention}, also discussed in Section~\ref{sec:Computational Efficiency}, improves training speed by optimizing memory usage; however, it still lags behind optimized matrix multiplication in terms of efficiency and faces challenges such as low GPU occupancy and unnecessary shared memory I/O operations. Therefore, future research should focus on accelerating computational efficiency at both the software and hardware levels while maintaining high performance.

\subsection{Efficient Fine-Tuning Design}
Section \ref{sec:Computational Efficiency} introduces a novel paradigm for parameter-efficient fine-tuning by fine-tuning attention weight matrices\cite{hu2021lora,dettmers2024qlora,kumari2023multi}. This approach enables the indirect training of large models with minimal parameters through low-rank decomposition, simulating parameter changes. However, when applied to diverse downstream generation tasks, fine-tuning only the self-attention and cross-attention layers often fails to meet performance requirements and can reduce effectiveness. Future research should explore how to achieve a balance between the number of trained parameters and generation performance by optimizing the fine-tuning of weight matrices.

\subsection{Interpretable Problems}
A substantial body of research has demonstrated that attention mechanisms are both computationally efficient and effective. On one hand, researchers strive to gain a deeper understanding of these mechanisms to optimize model performance. On the other hand, some scholars remain skeptical about their true effectiveness. Although attention mechanisms have been debated for their role in improving model interpretability, with careful design and the application of appropriate methods, they can indeed provide meaningful explanations in specific contexts. Future research should focus on how to better leverage attention mechanisms to enhance model transparency and interpretability, ultimately fostering greater understanding and trust in the model's predictions.

\subsection{3D Attention}
Attention layers play a crucial role in maintaining multi-view consistency in 3D generation and editing. The transition from 2D attention to 3D attention is expected to significantly impact the quality of 3D generation, making it essential to explore how attention mechanisms can be effectively applied or adapted in the 3D context. Although some efforts \cite{wang2024gscream,chen2024dge} have been made to incorporate attention into 3D generation, there remains substantial room for improvement, especially when handling complex backgrounds or environments. In the future, more researchers are likely to investigate 3D attention mechanisms to enhance the consistency and quality of generated 3D content.

\subsection{Applications and Challenges of Future Generative Diffusion Models}
Currently, most generative tasks based on diffusion models focus on single-task or single-modality research. However, in the future, generative models should move beyond focusing on specific tasks or domains. Instead, they should be capable of learning a wide range of tasks and knowledge through a unified architecture and training approach, making them more generalizable and adaptable. To achieve this, future research should introduce more sophisticated cross-modal attention mechanisms, enabling models to learn deeper semantic associations between different modalities. In parallel, efforts should be made to compress and simplify models so they can run efficiently on end-to-end devices such as mobile and embedded systems. Developing more efficient sampling methods to reduce both the number of generation steps and the computational cost at each step will also be essential. Furthermore, improving the interpretability and controllability of these models will enhance user understanding and experience. These advancements will pave the way for broad AI applications across fields such as healthcare, education, environmental protection, and scientific research, ultimately promoting social progress and human welfare.

\bibliographystyle{IEEEtran}
\bibliography{main}
\end{document}